\documentclass{article}

\usepackage{amsmath,amssymb,amsthm}
\usepackage{times}
\usepackage{xr}
\usepackage{setspace}
\usepackage{hyperref}
\usepackage{url}
\usepackage{mathtools}
\usepackage{graphicx}
\usepackage{algorithm,algorithmic}
\usepackage{natbib}
\newtheorem{lemma}{Lemma}
\newtheorem{theorem}{Theorem}
\newtheorem{proposition}{Proposition}
\newtheorem{assumption}{Assumption}

\newcommand\vecnorm[2]{\left|\!\left| #1 \right|\!\right|_{#2}}
\newcommand\vecindex[2]{#1#2}
\newcommand\tdvecindex[2]{#1;#2}
\newcommand\ndvecindex[2]{#1;\ldots ;#2}

\def \bal#1\eal{\begin{align}#1\end{align}}
\def \bals#1\eals{\begin{align*}#1\end{align*}}

\def\grad{\nabla}

\def\real{\mathbb{R}}

\def\P{\mathbb{P}}
\def\X{\mathcal{X}}

\title{Vector-Space Markov Random Fields via Exponential Families}
\date{\vspace{-5ex}}

\author{
Wesley Tansey\footnote{Department of Computer Science, University of Texas at Austin, \texttt{tansey@cs.utexas.edu} (corresponding author)} \\
Oscar Hernan Madrid Padilla\footnote{Department of Statistics and Data Sciences; University of Texas at Austin, \texttt{oscar.madrid@utexas.edu}} \\
Arun Sai Suggala\footnote{Department of Computer Science, University of Texas at Austin, \texttt{arunsai@utexas.edu}} \\
Pradeep Ravikumar\footnote{Department of Computer Science, University of Texas at Austin, \texttt{pradeepr@cs.utexas.edu}} \\
}

\begin{document}

\maketitle

\begin{abstract} 
We present Vector-Space Markov Random Fields (VS-MRFs), a novel class of undirected graphical models where each variable can belong to an arbitrary vector space. VS-MRFs generalize a recent line of work on scalar-valued, uni-parameter exponential family and mixed graphical models, thereby greatly broadening the class of exponential families available (e.g., allowing multinomial and Dirichlet distributions). Specifically, VS-MRFs are the joint graphical model distributions where the node-conditional distributions belong to generic exponential families with general vector space domains. We also present a sparsistent $M$-estimator for learning our class of MRFs that recovers the correct set of edges with high probability. We validate our approach via a set of synthetic data experiments as well as a real-world case study of over four million foods from the popular diet tracking app MyFitnessPal. Our results demonstrate that our algorithm performs well empirically and that VS-MRFs are capable of capturing and highlighting interesting structure in complex, real-world data. All code for our algorithm is open source and publicly available.

\end{abstract}

\section{Introduction}
\label{sec:introduction}

Undirected graphical models, also known as Markov Random Fields (MRFs), are a popular class of models for probability distributions over random vectors. Popular parametric instances include Gaussian MRFs, Ising, and Potts models, but these are all suited to specific data-types: Ising models for binary data, Gaussian MRFs for thin-tailed continuous data, and so on. Conversely, when there is prior knowledge of the graph structure but limited information otherwise, nonparametric approaches are available \cite{sudderth:etal:2010}. A recent line of work has considered the challenge of specifying classes of MRFs targeted to the data-types in the given application, when the structure is unknown. For the specific case of homogeneous data, where each variable in the random vector has the same data-type, \cite{yang:etal:2012} proposed a general subclass of homogeneous MRFs. In their construction, they imposed the restriction that each variable conditioned on other variables belong to a shared exponential family distribution, and then performed a Hammersley-Clifford-like analysis to derive the corresponding joint graphical model distribution, consistent with these node-conditional distributions. As they showed, even classical instances belong to this sub-class of MRFs; for instance, with Gaussian MRFs and Ising models, the node-conditional distributions follow univariate Gaussian and Bernoulli distributions respectively. 

\citet{yang:etal:2014} then proposed a class of \emph{mixed MRFs} that extended this construction to allow for random vectors with variables belonging to different data types, and allowing each node-conditional distribution to be drawn from a different univariate, uni-parameter exponential family member (such as a Gaussian with known variance or a Bernoulli distribution). This flexibility in allowing for different univariate exponential family distributions yielded a class of mixed MRFs over heterogeneous random vectors that were capable of modeling a much wider class of distributions than was previously feasible, opening up an entirely new suite of possible applications. 

To summarize, the state of the art can specify MRFs over heterogeneous data-typed random vectors, under the restriction that each variable conditioned on others belong to a uni-parameter, univariate exponential family distribution. But in many applications, such a restriction would be too onerous. For instance, a discrete random variable is best modeled by a categorical distribution, but this is a \emph{multi-parameter} exponential family distribution, and does not satisfy the required restriction above. Other multi-parameter exponential family distributions popular in machine learning include gamma distributions with unknown shape parameter and Gaussian distributions with unknown variance. Another restriction above is that the variables be scalar-valued; but in many applications the random variables could belong to more general vector spaces, for example a Dirichlet distribution.

As modern data modeling requirements evolve, extending MRFs beyond such restrictive paradigms is becoming increasingly important. In this paper, we thus extend the above line of work in \cite{yang:etal:2012,yang:etal:2014}. As opposed to other approaches which merely cluster scalar variables \cite{vats:etal:2012}, we allow node-conditional distributions to belong to a generic exponential family with a general vector space domain. We then perform a subtler Hammersley-Clifford-like analysis to derive a novel class of vector-space MRFs (VS-MRFs) as joint distributions consistent with these node-conditional distributions. This class of VS-MRFs provides support for the many modelling requirements outlined above, and could thus greatly expand the potential applicability of MRFs to new scientific analyses.

We also introduce an $M$-estimator for learning this class of VS-MRFs based on the sparse group lasso, and show that it is sparsistent, and that it succeeds in recovering the underlying edges of the graphical model. To solve the $M$-estimation problem, we also provide a scalable optimization algorithm based on Alternating Direction Method of Multipliers (ADMM) \cite{boyd:etal:2011}. We validate our approach empirically via synthetic experiments measuring performance across a variety of scenarios. We also demonstrate the usefulness of VS-MRFs by modeling a real-world dataset of over four million foods from the MyFitnessPal food database.

The remainder of this paper is organized as follows. Section \ref{sec:scalar_case} provides background on mixed MRFs in the uni-parameter, univariate case. Section \ref{sec:vector_case} details our generalization of the mixed MRF derivations to the vector-space case. Section \ref{sec:learning} introduces our $M$-estimator and derives its sparsistency statistical guarantees. Section \ref{sec:experiments} contains our synthetic experiments and the MyFitnessPal case study. Finally, Section \ref{sec:conclusion} presents concluding remarks and potential future work.

\section{Background: Scalar Mixed Graphical Models}
\label{sec:scalar_case}


Let $X = (X_1,X_2,\cdots X_p)$ be a $p$-dimensional random vector, where each variable $X_r$ has domain $\mathcal{X}_r$. An undirected graphical model or a Markov Random Field (MRF) is a family of joint distributions over the random vector $X$ that is specified by a graph $G = (V,E)$, with nodes corresponding to each of the $p$ random variables $\{X_r\}_{r=1}^{p}$, and edges that specify the factorization of the joint as:
\begin{align*}
	\mathbb{P}(X) \propto \prod_{C \in \mathcal{C}(G)} \, \psi_{C}(X_C),
\end{align*}
where $\mathcal{C}(G)$ is the set of fully connected subgraphs (or cliques) of the graph $G$, $X_C = \{X_{s}\}_{s \in C}$ denotes the subset of variables in the subset $C \subseteq V$, and $\{\psi_{C}(X_C)\}_{C \in \mathcal{C}(G)}$ are \emph{clique-wise} functions, each of which is a ``local function'' in that it only depend on the variables in the corresponding clique, so that $\psi_{C}(X_C)$ only depends on the variable subset $X_C$.

Gaussian MRFs, Ising MRFs, etc. make particular parametric assumptions on these clique-wise functions, but a key question is whether there exists a more flexible specification of the form of these clique-wise functions that is targeted to the data-type and other characteristics of the random vector $X$. 

For the specific case where the variables are scalars, so that the domains $\mathcal{X}_r \subseteq \real$, in a line of work, \cite{yang:etal:2012, yang:etal:2014} used the following construction to derive a subclass of MRFs targeted to the random vector $X$. Suppose that for variables $X_r \in \mathcal{X}_r$, the following (single-parameter) univariate exponential family distribution $P(X_r) = \exp \{ \theta_r B_r(X_r) + C_r(X_r) - A_r(\theta_r) \}$, 
with natural parameter scalar $\theta$, sufficient statistic scalar $B_r(X_r)$, base measure $C_r(X_r)$ and log normalization constant $A_r(\theta)$, serves as a suitable statistical model. Suppose that we use these univariate distributions to specify \emph{conditional} distributions:
\begin{equation}
\label{scalar_node_conditional}
\begin{array}{ll}
P(X_r | X_{-r}) = \exp \{ & E_r(X_{-r}) B_r(X_{r}) + \\
& C_r(X_r) - A_r(X_{-r}) \}
\end{array}\, ,
\end{equation}
where $E_r(\cdot)$ is an arbitrary function of the rest of the variables $X_{-r}$ that serves as the natural parameter. Would these node-conditional distributions for each node $r \in V$ be consistent with some joint distribution for some specification of these functions $\{E_r(\cdot)\}_{r \in V}$? Theorem 1 from \citet{yang:etal:2014} shows that there does exist a unique joint MRF distribution with the form: 
\begin{equation}
\label{scalar_joint}
\begin{array}{l}
 \hspace{-0.5in} P(X; \theta) = \exp \left\{ \displaystyle \sum_{r \in V} \theta_r B_r(X_r) \right. \\
  + \displaystyle \sum_{r \in V} \sum_{t \in N(r)} \hspace{-0.05in} \theta_{rt} B_t(X_t) B_r(X_r)
  + \dots \\
  + \hspace{-0.15in} \displaystyle \sum_{(t_1, \ldots, t_k) \in \mathcal{C}} \hspace{-0.15in} \theta_{t_1 \ldots t_k}(X) \prod_{j=1}^k B_{t_j}(X_{t_j})\\
  + \sum_{r \in V} C_r(X_r) - A(\theta) \left. \vphantom{\displaystyle \sum_{r \in V}} \right\} \end{array}\, ,
\end{equation}
where $A(\theta)$ is the log-normalization constant. Their proof followed an analysis similar to the Hammersley-Clifford Theorem~ \cite{lauritzen:1996}, and entailed showing that for a consistent joint, the only feasible conditional parameter functions $E_r(\cdot)$ had the following form:
\begin{equation}
\label{scalar_natural_parameters}
\hspace{-0.1in}\begin{array}{ll}
E_r(X_{-r}) =& \theta_r + \displaystyle \sum_{t \in N(r)} \theta_{rt} B_t(X_t) + \dots \\
             & + \displaystyle \sum_{t_2, \ldots, t_k \in N(r)} \theta_{r t_2 \ldots t_k}(X) \prod_{j=2}^k B_{t_j}(X_{t_j})
\end{array} \, ,
\end{equation}
where $\theta_{r \cdot} := \{ \theta_r, \theta_{rt}, \ldots, \theta_{r t_2 \ldots t_k} \}$ is a set of parameters, and $N(r)$ is the set of neighbors of node $r$.

While their construction allows the specification of targeted classes of graphical models for heterogeneous random vectors, the conditional distribution of each variable conditioned on the rest of the variables is assumed to be a single-parameter exponential family distribution with a scalar sufficient statistic and natural parameter. Furthermore, their Hammersley-Clifford type analysis and sparsistency proofs relied crucially on that assumption. However in the case of multi-parameter and multivariate distributions, the sufficient statistics are a vector; indeed the random variables need not be scalars at all but could belong to a more general vector space. Could one construct classes of MRFs for this more general, but prevalent, setting? In the next section, we answer in the affirmative, and present a generalization of mixed MRFs to the vector-space case, with support for more general exponential families.






\section{Generalization to the Vector-space Case}
\label{sec:vector_case}

Let $X = (X_1,X_2,\cdots X_p)$ be a $p$-dimensional random vector, where each variable $X_r$ belongs to a vector space $\mathcal{X}_r$. As in the scalar case, we will assume that a suitable statistical model for variables 
$X_r \in \mathcal{X}_r$ is an exponential family distribution \vspace{-0.1in}
\begin{equation}
\label{vector_exponential_family_distribution}
P(X_r) = \exp \{ \sum_{j=1}^{m_r} \theta_{\vecindex{r}{j}} B_{\vecindex{r}{j}}(X_r) + C_r(X_r) - A_r(\theta) \}, \vspace{-0.1in}
\end{equation}
with natural parameters $\{\theta_{\vecindex{r}{j}}\}_{j=1}^{m_r}$, and sufficient statistics
$\{B_{\vecindex{r}{j}}\}_{j=1}^{m_r}$, base measure $C_r(X_r)$ and log normalization constant $A_r(\theta)$. We assume the sufficient statistics $B_{\vecindex{r}{j}} : \mathcal{X}_r \mapsto \mathbb{R}$ lie in some Hilbert space $\mathcal{H}_{s}$, and moreover specify a minimal exponential family so that:\vspace{-0.1in}
\begin{equation}
\label{vector_theorem_minimal}
\displaystyle  \sum_{j=1}^{m_r} \alpha_j B_{\vecindex{r}{j}}(X_r) \neq c \, ,
\vspace{-0.1in}\end{equation}
for any constant $c$ and any vector $\alpha \neq \mathbf{0}$.  We note that even though the variables $\{X_r\}$ could lie in general vector spaces, the exponential family distribution above is finite-dimensional. However, it has multiple parameters, which is the other facet that distinguishes it from the single-parameter univariate setting of \cite{yang:etal:2012, yang:etal:2014}. We defer a generalization of our framework to infinite-dimensional exponential families to future work.

Suppose we use these general exponential family distributions to specify node-conditional distributions of variables $X_r$ conditioned on the rest of the random variables:
\begin{equation}
\label{vector_node_conditional}
\begin{array}{lll}
\hspace{-0.05in}\vspace{-0.03in}P(X_r | X_{-r}) &=& \exp \{ \sum_{j=1}^{m_r}  E_{\vecindex{r}{j}}(X_{-r}) B_{\vecindex{r}{j}}(X_{r}) \\
\\
&& \hspace{0.25in} + C_r(X_r) - A_r(X_{-r})\} \, ,
\end{array}
\end{equation}
where $\{E_{\vecindex{r}{j}}(X_{-r})\}_{j=1}^{m_r}$ are arbitrary functions of the rest of the variables that serve as natural parameters for the conditional distribution of $X_r$. As before, we ask the question whether these node-conditional distributions can be consistent with some joint distribution for some specification of the parameter functions $\{E_{\vecindex{r}{j}}(X_{-r})\}_{j=1}^{m_r}$; the following theorem addresses this very question.

\begin{theorem}
\label{thm:vector_theorem}
Let $X = (X_1,X_2,\ldots, X_p)$ be a p-dimensional random vector with node-conditional distribution of each random vector $X_r$ conditioned on the rest of random variables as defined in \eqref{vector_node_conditional}. These node-conditionals are consistent with a joint MRF distribution over the random vector $X$, that is, Markov with respect to a graph $G=(V, E)$ with clique-set $\mathcal{C}$, and with factors of size at most k, \textbf{if and only if} the functions $\{E_r()\}_{r\in V}$ specifying the node-conditional distributions have the form:\vspace{-0.1in}
\begin{equation}
\label{vector_natural_parameters}
\begin{split}
E_{\vecindex{r}{i}}(X_{-r}) = & \theta_{\vecindex{r}{i}} + \sum_{t \in N(r)} \sum_{j = 1}^{m_t} \theta_{\tdvecindex{\vecindex{r}{i}}{\vecindex{t}{j}}} B_{\vecindex{t}{j}}(X_t) + \dots \\
& \hspace{-0.5in} +  \sum_{\begin{subarray}{l}t_{2}, \dots, t_{k} \\ \in N(r)\end{subarray}} \sum_{\begin{subarray}{l}i_2 = 1\ldots m_{t_2}\\ \ldots \\ i_k = 1\ldots m_{t_k}\end{subarray}} \theta_{\ndvecindex{\vecindex{r}{i}}{\vecindex{t_k}{i_k}}} \prod_{j=2}^k B_{\vecindex{t_j}{i_j}}(X_{t_j})
\end{split}
\, ,
\end{equation}
where $\theta_{r\cdot} = \{\theta_{\vecindex{r}{i}}, \theta_{\tdvecindex{\vecindex{r}{i}}{\vecindex{t}{j}}}, \theta_{\ndvecindex{\vecindex{r}{i}}{\vecindex{t_k}{i_k}}}\}$ is a set of parameters, $m_t$ is the dimension of the sufficient statistic vector for the $t^{\text{th}}$ node-conditional distribution, and $N(r)$ is the set of neighbors of node $r$ in graph $G$. The corresponding consistent joint MRF distribution has the following form:\vspace{-0.1in}
\begin{equation}
\begin{array}{l}
P(X | \theta) = \exp \left\{ \displaystyle \sum_{r \in V} \displaystyle \sum_{i = 1}^{m_r} \theta_{\vecindex{r}{i}} B_{\vecindex{r}{i}}(X_r) + \dots \right.\\
                      + \displaystyle \sum_{\begin{subarray}{l}t_{1}, \dots, t_{k}  \in C \end{subarray}} \sum_{\begin{subarray}{l}i_1 = 1\ldots m_{t_1}\\ \ldots \\ i_k = 1\ldots m_{t_k}\end{subarray}} \theta_{\ndvecindex{\vecindex{t_1}{i_1}}{\vecindex{t_k}{i_k}}} \prod_{j=1}^k B_{\vecindex{t_j}{i_j}}(X_{t_j}) \\
                      + \left. \displaystyle \sum_{r \in V} C_r(X_r) - A(\theta) \right\}
\end{array}
\, .
\end{equation}
\end{theorem}
We provide a Hammersley-Clifford type analysis as proof of this theorem in the supplementary material, which however has subtleties not present in \cite{yang:etal:2012, yang:etal:2014}, due to the arbitrary vector space domain of $X_r$, and the multiple parameters in the exponential families, which consequently entailed leveraging the geometry of the corresponding Hilbert spaces $\{\mathcal{H}_{s}\{s \in V\}$ underlying the sufficient statistics $\{B_{sj}\}$.

The above Theorem \ref{thm:vector_theorem} provides us with a general class of vector-space MRFs (VS-MRFs), where each variable could belong to more general vector space domains, and whose conditional distributions are specified by more general finite-dimensional exponential families. Consequently, many common distributions can be incorporated into VS-MRFs that were previously unsupported or lacking in \cite{yang:etal:2012, yang:etal:2014}. For instance, gamma and Gaussian nodes, though univariate, require vector-space parameters in order to be fully modeled. Additionally, multivariate distributions that were impossible to use with previous methods, such as the multinomial and Dirichlet distributions are now also available. 

\vspace{-0.1in}


\subsection{Pairwise conditional and joint distributions}
\label{sec:vector_conditional_and_joint}

Given the form of natural parameters in \eqref{vector_natural_parameters}, the conditional distribution of a node $X_r$ given all other nodes $X_{-r}$ for the special case of pairwise MRFs (i.e. $k=2$) has the form
\begin{equation}
\label{vector_conditional_distribution_k2}
\hspace{-0.05in}\begin{array}{l}
P(X_r | X_{-r}, \theta_r, \theta_{rt}) = \exp \left\{ \vphantom{\displaystyle \sum_{t \in N(r)}}
                            \displaystyle \sum_{i = 1}^{m_r} \theta_{\vecindex{r}{i}} B_{\vecindex{r}{i}}(X_r) \right.\\
                            + \displaystyle \sum_{t \in N(r)} \displaystyle \sum_{i = 1}^{m_r} \sum_{j = 1}^{m_t} \theta_{\tdvecindex{\vecindex{r}{i}}{\vecindex{t}{j}}} B_{\vecindex{t}{j}}(X_t) B_{\vecindex{r}{i}}(X_r) \\
                             + C_r(X_r) - A_r(X_{-r}, \theta_{r\cdot})\left. \vphantom{\displaystyle \sum_{t \in N(r)}} \hspace{-0.05in}\right\} \\

                = \exp \left\{ \vphantom{\displaystyle \sum_{t \in N(r)}} \right.
                    \left\langle B_{r}(X_r), \theta_{r}  + \displaystyle \sum_{t \in N(r)} \theta_{rt} B_{t}(X_t) \right\rangle \\
                    + C_r(X_r) - A_r\left(\theta_{r} + \displaystyle \sum_{t \in N(r)} \theta_{rt}B_t(X_t)\right) \left. \vphantom{\displaystyle \sum_{t \in N(r)}} \right\}
\end{array}
\, ,
\end{equation}
where $\theta_r$ is a vector formed from scalars $\{\theta_{\vecindex{r}{i}}\}_{i = 1}^{m_r}$, $\theta_{rt}$ is a matrix of dimension $m_r \times m_t$ obtained from scalars $\theta_{\tdvecindex{\vecindex{r}{i}}{\vecindex{t}{j}}}$ and $\langle.,.\rangle$ represents dot product between two vectors. Thus, the joint distribution has the form
\begin{equation}
\label{vector_joint_distribution}
\begin{array}{l}
P(X | \theta) =\\ \exp \left\{ \displaystyle \sum_{r \in V} \left\langle B_{r}(X_r), \theta_{r}  + \displaystyle \sum_{t \in N(r)} \theta_{rt} B_{t}(X_t) \right\rangle \right. \\
                      + \left. \displaystyle \sum_{r \in V} C_r(X_r) - A(\theta) \right\}
\end{array}
\, ,
\end{equation}
with the log-normalization constant $A(\theta) = \log \int_{\mathcal{X}} \mathbf{\exp} \{ \sum_{r \in V} \langle B_{r}(X_r), \theta_{r}  + \sum_{t \in N(r)} \theta_{rt} B_{t}(X_t)\rangle + \sum_{r \in V} C_r(X_r)\}$. Since $A(\theta)$ is generally intractable to calculate, we next present an efficient approach to learning the structure of VS-MRFs.

\section{Learning VS-MRFs}
\label{sec:learning}

To avoid calculation of the log-normalization constant, we approximate the joint distribution in (\ref{vector_joint_distribution}) with the independent product of node conditionals, also known as the \textit{pseudo-likelihood},
\begin{equation}
\label{pseudo_likelihood}
P(X | \theta) \approx \prod_r P(X_r | X_{-r}, \theta_r, \theta_{rt}) \, .
\end{equation}
Let $\theta_{r\cdot} = \{\theta_r, \theta_{\setminus r} \}$ be the set of parameters related to the node-conditional distribution of node r, where $\theta_{\setminus r} = \{\theta_{rt}\}_{t \in V\setminus r}$. Since $A_r()$ is convex for all exponential families \cite{wainwright:etal:2008}, this gives us a loss function that is convex in $\theta_{r\cdot}$:
\begin{equation}
\label{loss_function}
\hspace{-0.15in}\begin{array}{ll}
\ell(\theta_{r\cdot}; \mathcal{D}) =&\hspace{-0.1in} -\frac{1}{n}\displaystyle \sum_i^n \left( \left\langle B_{r}(X^{(i)}_r), \theta_{r}  + \displaystyle \sum_{t \in V\setminus r} \theta_{rt} B_{t}(X^{(i)}_t)\right\rangle \right.\\
&\hspace{0.25in}-\left. A_r\left(\theta_{r} + \displaystyle \sum_{t \in V\setminus r} \theta_{rt}B_t(X^{(i)}_t)\right)\right)
\end{array}
\, .
\end{equation}
We then seek to find a sparse solution in terms of both edges and individual parameters by employing the sparse group lasso regularization penalty \cite{friedman:etal:2010, simon:etal:2013}:
\begin{equation}
\label{regularization_penalty}
R(\theta_{r\cdot}) = \lambda_1 \sum_{t \in V\setminus r} \sqrt{\nu_{rt}} \vecnorm{\theta_{rt}}{2} + \lambda_2 \vecnorm{\theta_{\setminus r}}{1} \, ,
\end{equation}
where $\nu_{rt} = m_r \times m_t$ is the number of parameters in the \textit{pseudo-edge} from node $r$ to node $t$ (i.e., the edge $(r, t)$ in the $r^{th}$ node-conditional). This yields a collection of independent convex optimization problems, one for each node-conditional.

\begin{equation}
\label{optimization_problem}
\begin{aligned}
& \underset{\theta_{r\cdot}}{\text{minimize}}
& & 
\ell(\theta_{r\cdot}; \mathcal{D}) + R(\theta_{r\cdot})
\end{aligned}
\end{equation}

We next present an approach to solving this problem based on Alternating Direction Method of Multipliers (ADMM) \cite{boyd:etal:2011}. 

\subsection{Optimization Procedure}
\label{sec:optimization_procedure}

We first introduce a slack variable $z$ into (\ref{optimization_problem}) to adhere to the canonical form of ADMM. For notational simplicity, we omit the data parameter $\mathcal{D}$ from the loss function and the subscripts in $\theta_{r\cdot}$ and $A_r$ since it is clear we are dealing with the optimization of a single node-conditional.
\begin{equation}
\begin{aligned}
& \underset{\theta}{\text{minimize}}
& & 
\ell(\theta) + R(z) \\
& \text{subject to}
& & \theta = z 
\end{aligned}
\hspace{0.2in}\, ,
\end{equation}
where $length(\theta) = \tau$. The augmented Lagrangian is
\begin{equation}
\label{eqn:augmented_lagrangian}
L_{\alpha}(\theta, z, \rho) = \ell(\theta) + R(z) + \rho^T(\theta - z) + (\alpha / 2) \vecnorm{\theta - z}{2}^2 \, .
\end{equation}

Defining the residual of the slack $r = \theta - z$, we instead use the scaled form with $u = (1/\alpha)\rho$. ADMM proceeds in an alternating fashion, performing the following updates at each iteration:
\begin{align}
\label{admm_updates_theta}
\hspace{-0.12in}\theta^{k+1}& = \underset{\theta}{\text{argmin}} \left(\ell(\theta) + (\alpha / 2)\vecnorm{\theta - z^{k} + u^{k}}{2}^2  \right) \\
\label{admm_updates_z}
\hspace{-0.12in}z^{k+1}& = \underset{z}{\text{argmin}} \left(R(z) + (\alpha / 2)\vecnorm{\theta^{k+1} - z + u^{k}}{2}^2  \right) \\
\label{admm_updates_u}
\hspace{-0.12in}u^{k+1}& = u^k + \theta^{k+1} - z^{k+1}
\end{align}

\paragraph{Updating $\theta^{k+1}$.} The $j^{th}$ subgradient of $\theta$ is $g_j(\theta) = -\overline{B}_j + \grad_j \overline{A}(\theta) + \alpha (\theta_j + z^k_j - u^k_j)$. Note that the log-partition function, $A(\eta)$, over the natural parameters, $\eta = B\theta$, is available in closed form for most commonly-used exponential families. Thus, $\grad^2 \overline{A}(\theta)$ is a weighted sum of rank-one matrices. In cases where the number of samples is much less than the total number of parameters (i.e. $n << \tau$), we can efficiently calculate an exact Newton update in $\mathcal{O}(\tau)$ by leveraging the matrix inversion lemma \cite{boyd:etal:2009}. Otherwise, we use a diagonal approximation of the Hessian and perform a quasi-Newton update.

\paragraph{Updating $z^{k+1}$.} We can reformulate (\ref{admm_updates_z}) as the \textit{proximal operator} \cite{parikh:etal:2013} of $R(z)$:
\begin{equation}
\vspace{-0.1in}\text{\textbf{prox}}_{R/\alpha}(y) = \underset{z}{\text{argmin}} \left(R(z) + (\alpha / 2)\vecnorm{z - y}{2}^2 \right) \, ,
\label{z_proximal_operator}
\end{equation}

where $y = \theta^{k+1} + u^k$. From \citet{friedman:etal:2010}, it is straightforward to show that the update has a closed-form solution for each $j^{th}$ block of edge parameters,
\begin{equation}
\label{z_closed_form}
z^{k+1}_j = \frac{ \left( \vecnorm{S(\alpha (y_j), \lambda_2)}{2} - \sqrt{\nu_j}\lambda_1 \right)_+ S(\alpha(y_j), \lambda_2)}{\alpha\vecnorm{S(\alpha(y_j), \lambda_2)}{2} + \sqrt{\nu_j}\lambda_1 (1 - \alpha)} \, ,
\end{equation}
where $S(x, \lambda)$ is the soft-thresholding operator on $x$ with cutoff at $\lambda$.

\paragraph{Updating $u^{k+1}$.} Per ADMM, closed-form is given in (\ref{admm_updates_u}).

We iterate each of the above update steps in turn until convergence, then \texttt{AND} pseudo-edges when stitching the graph back together.

\subsection{Domain constraints}
\label{sec:domain_constraints}

Many exponential family distributions require parameters with bounded domain. These bounds correspond to affine constraints on subsets of $\theta$ in the ADMM algorithm.\footnote{Note that these subsets are different than the edge-wise groups that are $L_2$-penalized. Rather, these constraints apply to the sum of the $i^{th}$ value of each edge parameter and the $i^{th}$ bias weight.} Often these constraints are simple implicit restrictions to $\real^+$ or $\real^-$. In these cases the log-normalization function $A(\eta)$ serves as a built-in \textit{log-barrier} function. For instance, a normal distribution with unknown mean $\mu$ and unknown variance $\sigma^2$ has natural parameters $\eta_1 = \frac{\mu}{\sigma^2}$ and $\eta_2 = -\frac{1}{2\sigma^2}$, implying $\eta_2 < 0$. However, since $A(\eta) = -\frac{\eta_1^2}{4\eta_2} - \frac{1}{2}\ln (-2\eta_2)$, this constraint will be effectively enforced so long as we are given a feasible starting point for $\eta$. Such a feasible point can always be discovered using a standard phase I method \cite{boyd:etal:2009}. In the case of equality requirements, such as categorical and multinomial distributions, we can directly incorporate the constraints into the ADMM algorithm and solve an equality-constrained Newton's method when updating $\theta$.



\subsection{Sparsistency}
\label{sec:sparsistency}

We next provide the mathematical conditions that ensure with high probability our learning procedure recovers the true graph structure underlying the joint distribution. Our results rely on similar sufficient conditions to those imposed in papers analyzing the Lasso \cite{wainwright:2009} and the $l_1/l_2$ penalty in \cite{Jalali:etal:2011}. Before stating the assumptions, we introduce the notation used in the proof.

\subsubsection{Notation}
Let $N(r) = \{t: \theta^*_{rt} \neq 0\}$  be the  true neighbourhood of node \textit{r} and let $d_r$ be the degree of r, \textit{i.e,} $d_r = |N(r)|$. And $S_r$ be the index set of parameters $\{\theta^*_{\tdvecindex{\vecindex{r}{j}}{\vecindex{t}{k}}} : t \in N(r)\}$ and similarly $S_r^c$ be the index of parameters  $\{\theta^*_{\tdvecindex{\vecindex{r}{j}}{\vecindex{t}{k}}} : t \notin N(r)\}$. From now on we will overload the notation and simply use $S$ and $S^c$ instead of $S_r$ and $S_r^c$. Let $S_r^{(ex)} = \{\theta^*_{\tdvecindex{\vecindex{r}{j}}{\vecindex{t}{k}}} : \theta^*_{\tdvecindex{\vecindex{r}{j}}{\vecindex{t}{k}}} \neq 0\  \land \ t \in N(r)\}$.

Let $Q_r^n = \nabla^2 \ell(\theta^*_{r\cdot}; \mathcal{D})$ be the sample Fischer Information matrix at node r. As before, we will ignore subscript $r$ and use $Q^n$ instead of $Q_r^n$. Finally,  we write $Q_{SS}^n$ for the sub-matrix indexed by $S$. 

We use the group structured norms defined in \cite{Jalali:etal:2011} in our analysis. The group structured norm $\vecnorm{u}{\mathcal{G},a,b}$ of a vector $u$ with respect to a set of disjoint groups $\mathcal{G} = \{G_1,\ldots, G_T\}$ is defined as $\vecnorm{(\vecnorm{u_{G_1}}{b}, \ldots, \vecnorm{u_{G_T}}{b})}{a}$. We ignore the group $\mathcal{G}$ and simply use $\vecnorm{u}{a,b}$ when it is clear from the context. Similarly the group structured norm $\vecnorm{M}{(a,b),(c,d)}$ of a matrix $M_{p \times p}$ is defined as 
$\vecnorm{(\vecnorm{M^1}{c,d}, \ldots, \vecnorm{M^p}{c,d})}{a,b}$. In our analysis we always use $b = 2$, $d = 2$ and to minimize the notation we use $\vecnorm{M}{a,c}$ to denote $\vecnorm{M}{(a,2),(c,2)}$. And we define $\vecnorm{M}{max}$ as $\underset{i,j}{max}\ |M_{i,j}|$, i.e, element wise maximum of M.

\subsubsection{Assumptions}
\label{assumptions}
Let us begin by imposing assumptions on the sample Fisher Information matrix $Q^n$.
\begin{assumption}
\label{dependency_assumption}
\textit{Dependency condition:} $\Lambda_{min}(Q_{SS}^n) \geq C_{min}$.
\end{assumption}
\begin{assumption}
\label{mutual_incoherence_bound}
\textit{Incoherence condition:} \\
$\vecnorm{Q_{S^cS}^n (Q_{SS}^n)^{-1}}{\infty,2} \leq \frac{m_{min}}{m_{max}}\frac{(1 -\alpha)}{\sqrt{d_r}}$ for some $\alpha \in (0,1]$ , where $m_{max}$ $=$ $\underset{t }{\text{max}}\ m_t$, $m_{min}$ $=$ $\underset{t }{\text{min}}\ m_t$.
\end{assumption}
\begin{assumption}
\textit{Boundedness:}\\ 
$\Lambda_{max}(E[ B\left(X_{V\setminus r}\right)B\left(X_{V\setminus r}\right)^T ]) \leq D_{max} < \infty$, where $B\left(X_{V\setminus r}\right)$ is a vector such that $B\left(X_{V\setminus r}\right) =  \{B_{t}(X_t)\}_{t \in V\setminus r}$.
\end{assumption}

Note that the sufficient statistics \{$B_{\vecindex{r}{i}}(X_r)\}_{i = 1}^{m_r}$ of node $r$ need not be bounded. So to analyze the M-estimation problem, we make the following assumptions on log-partition functions of joint and node-conditional distributions. These are similar to the conditions imposed for sparsistency analysis of GLMs.
\begin{assumption}
\label{log_partition_joint_assumption}
The log partition function of the joint distribution satisfies the following conditions:  for all $r \in V$ and $i \in [m_r]$
\begin{enumerate}
\item there exists constants $k_m, k_v$ such that $E[B_{\vecindex{r}{i}}(X_r)] \leq k_m$ and $E[B_{\vecindex{r}{i}}(X_r)^{2}] \leq k_v$,
\item there exists constant $k_h$ such that $max_{u:|u| \leq 1} \frac{\partial^2A(\theta)}{\partial \theta^2_{\vecindex{r}{i}}}(\theta^*_{\vecindex{r}{i}} + u, \theta^*_{r\cdot}) \leq k_h$,
\item for scalar variable $\eta$ , we define a function $\bar{A}_{r,i}$ as: 
\begin{equation}
\label{modified_log_partition}
\begin{array}{l}
\hspace{-0.15in}\bar{A}_{r,i}(\eta;\theta) =  \log \int_{\mathcal{X}_p}  \exp\Big\{\eta B_{\vecindex{r}{i}}(X_r)^2 + \sum_{s \in V} C_s(X_s)  \\
\hspace{-0.15in}+  \sum_{s \in V} \left\langle B_{s}(X_s), \theta_{s}  + \displaystyle \hspace{-0.1in} \sum_{t \in N(s)} \theta_{st} B_{t}(X_t)\right\rangle
\Big\}d(x)
\end{array}
\end{equation}
Then, there exists a constant $k_h$ such that $max_{u:|u| \leq 1} \frac{\partial^2A_{r,i}(\eta;\theta^*_{r\cdot})}{\partial \eta^2}(u) \leq k_h$.
\end{enumerate}
\end{assumption}
\begin{assumption}
\label{node_conditional_assumption}
For all $r \in V$, the log-partition function $A_r(.)$ of the node wise conditional distribution satisfy that there exists functions  $k_{1}\left(n,p\right)$ and $k_{2}\left(n,p\right)$   such that for all feasible pairs $\theta$  and $X$, $\vecnorm{\nabla^{2}A_r\left(a\right)}{max}$ $\leq$  $k_{1}\left(n,p\right)$  where $a$ $\in$  $\left[b,b+4\, k_{2}\left(n,p\right)\max\left\{ \log\left(n\right),\log\left(p\right)\right\} \mathbf{1}\right]$  for  $b$  $:=$  $\theta_{r}+ \sum_{t \in V\setminus r} \theta_{rt} B_{t}(X_t)$, where for vectors $u$ and $v$ we define $[u,v]:= \otimes_{i} [u_i,v_i]$.  Moreoever, we assume that $\vecnorm{\nabla^{3}A_r\left(b\right)}{max}$ $\leq$ $k_{3}\left(n,p\right)$ for all feasible pairs $X$  and $\theta$.
\end{assumption}

\subsubsection{Sparsistency Theorem}
Given these assumptions in \ref{assumptions} we are now ready to state our main sparsistency result.

\begin{theorem}
\label{sparsistency_theorem_statement}
Consider the vector space graphical model distribution in \eqref{vector_joint_distribution} with true parameters $\theta^*$, edge set $E$  and vertex set $V$ such that the assumptions \ref{dependency_assumption}-\ref{node_conditional_assumption} hold. Suppose that $\theta^*$ satisfies $\underset{(r,t) \in E}{\text{min}} \vecnorm{\theta_{rt}^{*}}{2} \geq  \frac{10\ m_{max}}{C_{\text{min}}}\left(\lambda_{1}+\lambda_{2}\right)$  and regularization parameters $\lambda_{1}, \lambda_{2}$  satisfy  $M_{1}\,\frac{2-\alpha}{\alpha}\,\frac{m_{max}}{m_{min}}\,\sqrt{k_{1}\left(n,p\right)}\,\sqrt{\frac{\text{log}(pm_{max}^2)}{n}}$  $\leq$ $\lambda_{1}+\lambda_{2}$ $\leq$ $M_{2}\,\frac{2-\alpha}{\alpha}\,k_1\left(n,p\right)\,k_2\left(n,p\right)$ for positive constants $M_{1}$ and $M_{2}$ and $\lambda_2$ $<$ $\left(\frac{\alpha}{2 - \alpha + 2 \ {m_{max}}/{m_{min}}}\right)\, \lambda_1$. Then, there exists constants $L$, $c_1$, $c_2$ and $c_3$ such that if  $n \geq \text{max}\big\{L\,\frac{m_{max}^9}{m_{min}}d^2\,k_1\left(n,p\right)\,\left(k_3\left(n,p\right)\right)^2\,\left(\text{log}p^{\prime}\right)^2\text{log}\left(p\,m_{max}^2\right),\\ \frac{4\,\text{log}\left(p\,m_{max}^2\right)}{k_1\left(n,p\right)\,k_4\,k_2\left(n,p\right)^2},
\frac{8\,k_{h}^{2}}{k_{4}^{2}}\,\text{log}\left(\sum_{t} m_t\right)\big\} $, with probability at least $1- c\,\left(p^{\prime}\right)^{-3}\left(\sum_{t} m_t\right)- \text{exp}(-c_2\,n)-\text{exp}(-c_3\,n)$, the following statements hold.
\vspace{-0.1in}\begin{itemize}
\item For each node $r \in V$, the solution of the M-estimation problem \eqref{optimization_problem} is unique
\item Moreover, for each node $r \in V$ the M-estimation problem recovers the true neighbourhood exactly.
\end{itemize}
\vspace{-0.15in}where  $m_{max}$ $=$ $\underset{t }{\text{max}}\ m_t$, $m_{min}$ $=$ $\underset{t }{\text{min}}\ m_t$ , $p^{\prime}$ $=$ $\text{max}(n,p)$.
\end{theorem}
\vspace{-0.2in}The  proof of Theorem \ref{sparsistency_theorem_statement} follows along similar lines to the sparsistency proof in \cite{yang:etal:2014}, albeit with a subtler analysis to support general vector-spaces. It is based on the primal dual witness proof technique and relies on the previous results. We refer the interested reader to the supplementary material for additional details regarding the proofs.

%

\section{Experiments}
\label{sec:experiments}

We demonstrate the effectiveness of our algorithm on both synthetic data and a real-world dataset of over four million foods logged on the popular diet app, MyFitnessPal.


\vspace{-0.1in}\begin{figure}[tbh]
\begin{center}
\label{fig:synthetic_results}
\includegraphics[scale=0.3,trim=0 0.7in 0 0]{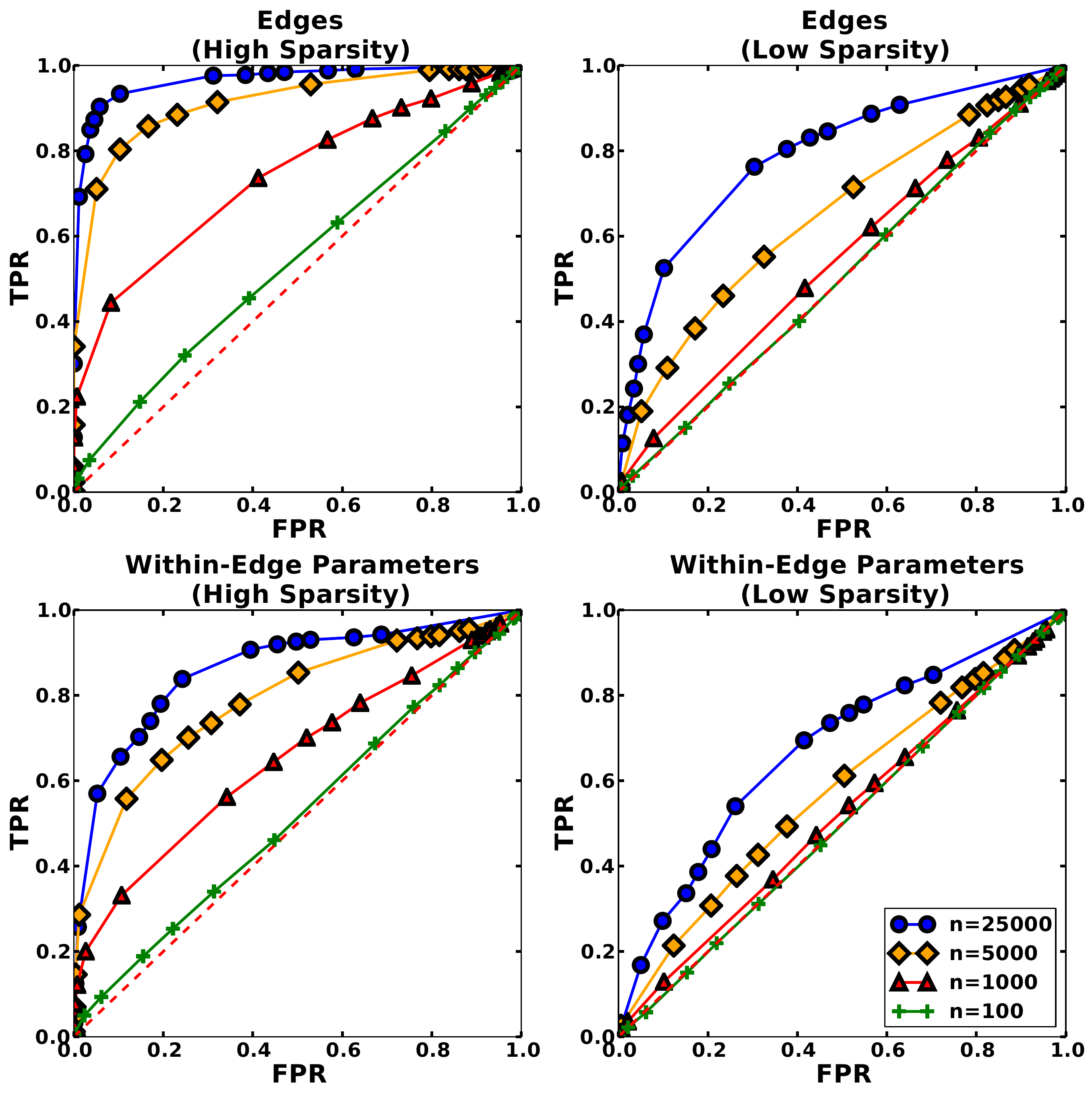}
\end{center}
\caption{ROC curves for our synthetic experiments. The top left and bottom left plots show both edge as well as within-edge-parameter recovery performance respectively, for graphs with a high degree of sparsity. The two right plots show the same performance measures, but for graphs with a relatively low degree of sparsity. The low sparsity scenario is more challenging, requiring more data to recover the majority of the graph.\vspace{-0.2in}}
\end{figure}

\subsection{Synthetic experiments}
The synthetic experiments were run on a vector-space mixed MRF consisting of eight Bernoulli, eight gamma (with unknown shape and rate), eight Gaussian (with unknown mean and variance), and eight Dirichlet (k=3) nodes. The choice of these node-conditional distributions is meant to highlight the ability of VS-MRFs to model many different types of distributions. Specifically, the Bernoulli represents a univariate, uni-parameter distribution that would still be possible to incorporate into existing mixed models. The gamma and Gaussian distributions are both multi-parameter, univariate distributions which would have required fixing one parameter (e.g. fixing the Gaussians' variances) to be compatible with previous approaches. Finally, the Dirichlet distribution is multi-parameter \textit{and} multivariate, thereby making VS-MRFs truly unique in their ability to model this joint distribution.

For each experiment, we conducted 30 independent trials by generating random weights and sampling via Gibbs sampling with a burn-in of 2000 and thinning step size of 10. We consider two different sparsity scenarios: high (90\% edge sparsity, 50\% intra-edge parameter sparsity) and low (50\% edge sparsity, 10\% intra-edge parameter sparsity). Edge recovery capability is examined by fixing $\lambda_2$ to a small value and varying $\lambda_1$ over a grid of values in the range $[0.0001, 0.5]$; parameter recovery is examined analogously by fixing $\lambda_1$ and varying $\lambda_2$. We use \texttt{AND} graph stitching and measure the true positive rate (TPR) and false positive rate (FPR) as the number of samples increases from 100 to 25K.

Figure \ref{fig:synthetic_results} shows the ROC curves at both the edge and parameter levels. The results demonstrate that our algorithm improves well as the dataset size scales. They also illustrate that graphs with a higher degree of sparsity are easier to recover with fewer samples. In both the high and low sparsity graphs, the algorithm is better able to recover the coarse-grained edge structure than the more fine-grained within-edge parameter structure, though both improve favourably with the size of the data.

\begin{figure*}[tbh]
\begin{center}
\includegraphics[scale=0.06,trim=0 4in 0 2in]{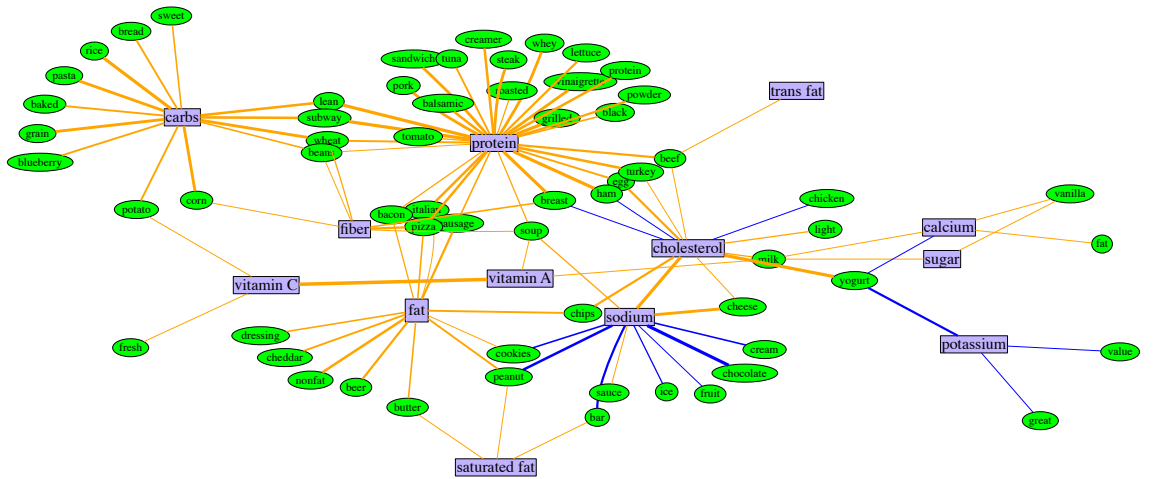}
\end{center}
\caption{\label{fig:mfp_top} The top 100 edges in the MyFitnessPal food graph. Purple rectangular nodes correspond to macro- and micro-nutrients; green oval nodes correspond to food description terms. Edge color is determined by the approximate effect of the edge on the means of the node-conditionals: darker, blue edges represent lower means; brighter, orange edges represent higher means; thickness corresponds to the norm of the edge weight.\vspace{-0.2in}}
\end{figure*}

\subsection{MyFitnessPal Food Dataset}
MyFitnessPal\footnote{\url{http://myfitnesspal.com}} (MFP) is one of the largest diet-tracking apps in the world, with over 80M users worldwide. MFP has a vast crowd-sourced database of food data, where each food entry contains a description, such as ``Trader Joe's Organic Carrots,'' and a vector of sixteen macro- and micro-nutrients, such as fat and vitamin C. 

We treat these foods entries as random vectors with an underlying VS-MRF distribution, which we learn treating the food entries in the database as samples from the underlying VS-MRF distribution. The text descriptions are tokenized, resulting in a dictionary of approximately 2650 words; we use a Bernoulli distribution to model the conditional distribution of each word. The conditional distribution of each nutrient (on a per-calorie basis) is generally gamma distributed, but contains spikes at zero\footnote{This is common in foods since many dishes are marketed as ``fat free'' or contain low nutrient density (e.g. soda).} and large outlier values.\footnote{This occurs when foods contain few calories but a large amount of some micro-nutrient (e.g. multi-vitamins)} The gamma distribution is undefined at zero, and the outlier values can result in numerical instability during learning, which thus suggests using a distribution other than the vanilla gamma distribution. Such zero-inflated data are common in many biostatistics applications, and are typically modeled via a mixture model density of the form $p(Z) = \pi \, \delta_0 + (1-\pi)\, g(z)$, where $\delta_0$ is the dirac delta at zero, and $g(z)$ is the density of the non-zero-valued data. Unfortunately, such mixture models are not generally representable as exponential families. 

To overcome this, we introduce the following class of \textbf{point-inflated exponential family} distributions. For any random variable $Z \in \mathcal{Z}$, consider any exponential family $P(Z) = \exp(\eta^T B(Z) + C(Z) - A(\eta))$, with sufficient statistics $B(\cdot)$, base measure $C(\cdot)$, and log-normalization constant $A(\cdot)$. We consider an inflated variant of this random variable, inflated at some value $j$; note that this could potentially lie outside the domain $\mathcal{Z}$, in which case the domain of the inflated random variable would become $\mathcal{Z} \cup \{j\}$. We then define the corresponding point-inflated exponential family distribution as:
$$
P_{\text{infl}}(Z) = \exp \big\{ \eta_{0} \mathbb{I}(Z = j) + \eta_{1}^T B(z) + C(z) - A_{\text{infl}}(\eta) \big\} \, ,
$$
where $A_{\text{infl}}(\eta)$ is the log-normalization constant of the point-inflated distribution which can be expressed in terms of the log-normalization constant $A(\cdot)$ of the uninflated exponential family distribution: 
$A_{\text{infl}}(\eta) = \log\big(\exp\{\eta_0\} - \exp\{\eta_{1}^T B(j) \mathbb{I}(j \in \mathcal{Z})\} + \exp\{A(\eta_{1})\} \big).$
Thus, as long as we have a closed form  $A(\cdot)$ for the log-partition function of the base distribution, we can efficiently calculate $P_{\text{infl}}(Z)$. The definition also permits an arbitrary number of inflated points by recursively specifying the base distribution as another point-inflated model. We model each of the MFP nutrients via a two-point-inflated gamma distribution, with points at zero and a winsorized outlier bucket. 

Due to the size of the overall graph, presenting it in closer detail here is not feasible. To give a qualitative perspective of the relationships captured by our algorithm, we selected the top 100 edges in the MFP food graph by ranking the edges based on their L2-norm. We then calculated their approximate contribution to the mean of their corresponding node-conditionals to determine edge color and thickness. Figure \ref{fig:mfp_top} shows the results of this process, with edges that contribute positively colored in orange and edges that contribute negatively colored in blue; edge thickness corresponds to the magnitude of the contribution. A high-level view of the entire learned graph is available in the supplementary materials.

Several interesting relationships can be discovered, even from just this small subset of the overall model. For instance, the negative connection between ``peanut'' and sodium may seem counter-intuitive, given the popularity of salted nuts, yet on inspection of the raw database it appears that indeed many peanut-based foods are actually very low in sodium on a per-calorie basis. As another example, ``chips'' are often thought of as a high-carb food, but the graph suggests that they actually tend to be a bigger indicator of high fat. In general, we believe there is great potential for wide-ranging future uses of VS-MRFs in nutrition and other scientific fields, with the MFP case study only scratching the surface of what can be achieved.\vspace{-0.1in} 





\section{Conclusion}
\label{sec:conclusion}

We have presented vector-space MRFs as a flexible and scalable approach to modeling complex, heterogeneous data. In particular, we  generalize the concept of mixed MRFs to allow for node-conditional distributions to be distributed according to a generic exponential family distribution, that is potentially  multi-parameter and even multivariate. Our VS-MRF learning algorithm has reassuring sparsistency guarantees and was validated against a variety of synthetic experiments and a real-world case study. We believe that the broad applicability of VS-MRFs will make them a valuable addition to the scientific toolbox. All code for our VS-MRF implementation is publicly available.\footnote{\url{https://github.com/tansey/vsmrfs}}


\appendix

\section{Proof of Theorem 1}
\label{sec:main_theorem}
\label{sec:vector_proof}
The proof follows the same lines as the proof in \citet{yang:etal:2014}. Let us denote $Q(X)$ as $ \log{(P(X)/P(0))}$. Note that $X = (X_1,X_2,\cdots X_p)$ and each $X_r$ belongs to a vector space. Given any $X$, let us denote $\bar{X}_s$ as $\bar{X}_s = (X_1,\cdots,X_{s-1},0,X_{s+1},\cdots,X_p)$. Consider the following expansion for $Q(X)$:
\begin{equation}
\label{vector_proof_Q_expansion}
\begin{array}{l}
Q(X) = \\
\displaystyle \sum_{\begin{subarray}{l} t \in \{1,\cdots, p\} \end{subarray}} \mathcal{I}[X_t \neq 0]G_{t}(X_t) + \cdots \\
\\
 + \displaystyle \sum_{\begin{subarray}{l} t_1,\cdots t_k \in \\ \{1, \cdots, p\}\end{subarray}} \mathcal{I}[X_{t_1} \neq 0, \ldots X_{t_k} \neq 0] G_{t_1 \ldots t_k}(X_{t_1}\ldots X_{t_k})
\end{array}
\end{equation}
where $\mathcal{I}$ is the indicator function which takes value 1 if its argument evaluates to true and 0 otherwise.

Using some simple algebra and the definition $Q(X) = \log{(P(X)/P(0))}$ we can show that
\begin{equation}
\label{vector_proof_Q_definition2}
\begin{array}{l}
\exp{(Q(X) - Q(\bar{X_s}))} = \frac{P(X_s|X_1,\cdots,X_{s-1},X_{s+1},\cdots,X_p)}{P(0|X_1,\cdots,X_{s-1},X_{s+1},\cdots,X_p)}
\end{array}
\end{equation}

From \eqref{vector_proof_Q_expansion} we have the following:
\begin{equation}
\label{vector_proof_Q_expansion2}
\begin{array}{l}
(Q(X) - Q(\bar{X_s})) = \\
\mathcal{I}[X_s \neq 0]\left(G_s(X_s) + \displaystyle \sum_{t \in \{1, \cdots, p\}\setminus s}\hspace{-0.2in} \mathcal{I}[X_t \neq 0]G_{s,t}(X_s,X_t)\right.\\
\left. + \displaystyle \sum_{\begin{subarray}{l} t_2,\cdots t_k \in \\ \{1, \cdots, p\}\setminus s\end{subarray}}\hspace{-0.2in} \mathcal{I}[X_{t_2} \neq 0, \ldots X_{t_k} \neq 0] G_{s,t_2 \ldots t_k}(X_s,\ldots X_{t_k})\right)
\end{array}
\end{equation}

Since the node conditional distribution follows the exponential family distribution defined in \eqref{vsmrf-vector_node_conditional} we can show that:
\begin{equation}
\label{vector_proof_Q_expansion3}
\begin{array}{l}
\log \frac{P(X_s|X_1,\cdots,X_{s-1},X_{s+1},\cdots,X_p)}{P(0|X_1,\cdots,X_{s-1},X_{s+1},\cdots,X_p)} = \\
 \left\langle E_s(X_{-s}), B_s(X_{s}) - B_s(0) \right\rangle + \left(C_s(X_s) - C_s(0)\right)
\end{array}
\end{equation}

Using \eqref{vector_proof_Q_expansion2} and \eqref{vector_proof_Q_expansion3} for left and right hand sides of \eqref{vector_proof_Q_definition2} and setting $X_t$ = 0 for all $t \neq s$ we obtain:
\begin{equation*}
\label{vector_proof_final_equality1}
\begin{array}{lll}
\mathcal{I}[X_s \neq 0]G_s(X_s) &=& \left\langle E_s(0), B_s(X_s) - B_s(0)\right\rangle  \\
&&+(C_s(X_s) - C_s(0))
\end{array}
\end{equation*} 

Similarly setting $X_r$ = 0 for all $r \notin \{s,t\}$ we obtain:
\begin{equation*}
\label{vector_proof_final_equality2}
\begin{array}{l}
\mathcal{I}[X_s \neq 0]G_s(X_s) + \mathcal{I}[X_s \neq 0, X_t \neq 0]G_{s,t}(X_s,X_t) = \\
\left\langle E_s(0\cdots X_t \cdots,0), B_s(X_s) - B_s(0)\right\rangle + (C_s(X_s) - C_s(0))
\end{array}
\end{equation*} 

Similarly, replacing $X_s$ with $X_t$ in \eqref{vector_proof_Q_definition2} and setting $X_r$ = 0 for all $r \notin \{s,t\}$ we obtain:
\begin{equation*}
\label{vector_proof_final_equality3}
\begin{array}{l}
\mathcal{I}[X_t \neq 0]G_t(X_t) + \mathcal{I}[X_s \neq 0, X_t \neq 0]G_{s,t}(X_s,X_t) = \\
\left\langle E_t(0\cdots X_s \cdots,0), B_t(X_t) - B_t(0)\right\rangle + (C_t(X_t) - C_t(0))
\end{array}
\end{equation*} 
From the above three equations we arrive at the following equality:
\begin{equation}
\label{vector_theorem_equality0}
\begin{array}{l}
\left\langle E_s(0\cdots X_t \cdots,0) - E_s(0), B_s(X_s) - B_s(0)\right\rangle = \\
\left\langle E_t(0\cdots X_s \cdots,0) - E_t(0), B_t(X_t) - B_t(0)\right\rangle
\end{array}
\end{equation}
The above equality should hold for the node conditional distributions to be consistent with the joint MRF distribution over \textit{X} with respect to graph \textit{G}.
So we need to find the form of $E_r()$ that satisfies the above equation.
Omitting zero vectors for clarity from \eqref{vector_theorem_equality0}, we get the following:
\begin{equation}
\label{vector_theorem_equality}
\hspace{-0.1in}
\begin{array}{lll}
\left\langle E_{t}(X_s), B_{t}(X_t)\right\rangle & = & \left\langle E_{s}(X_t), B_{s}(X_s)\right\rangle \\
\\
\displaystyle \sum_j E_{\vecindex{t}{j}}(X_s) B_{\vecindex{t}{j}}(X_t) & = & \displaystyle \sum_l E_{\vecindex{s}{l}}(X_t) B_{\vecindex{s}{l}}(X_s) \\
\end{array}
\end{equation}

We rewrite the natural parameter functions as
\begin{equation}
\label{vector_theorem_basis_expansion2}
\begin{array}{lll}
E_{\vecindex{t}{j}}(X_s) & = & \displaystyle \sum_l \theta_{\tdvecindex{\vecindex{s}{l}}{\vecindex{t}{j}}} B_{\vecindex{s}{l}}(X_s) + \overline{B}_{\vecindex{t}{j}}(X_s) \\
\\
E_{\vecindex{s}{l}}(X_t) & = & \displaystyle \sum_j \overline{\theta}_{\tdvecindex{\vecindex{s}{l}}{\vecindex{t}{j}}} B_{\vecindex{t}{j}}(X_t) + \overline{B}_{\vecindex{s}{l}}(X_t)
\end{array}
\end{equation}
where  $\forall j$ $\overline{B}_{\vecindex{t}{j}}(X_s)$ are functions in the Hilbert space $\mathcal{H}_{s}$ orthogonal to the span of functions $B_s(X_s)$, and $\forall j$ $\overline{B}_{\vecindex{s}{l}}(X_t)$ are functions in the Hilbert space $\mathcal{H}_{t}$ orthogonal to the span of functions  $B_t(X_t)$; and $\theta_{\tdvecindex{\vecindex{s}{l}}{\vecindex{t}{j}}}$,  $\overline{\theta}_{\tdvecindex{\vecindex{s}{l}}{\vecindex{t}{j}}}$ are scalars.
Combining (\ref{vector_theorem_equality}) and (\ref{vector_theorem_basis_expansion2}), we get
\begin{equation}
\label{vector_theorem_expanded_equality}
\begin{split}
\displaystyle \sum_j \sum_l \theta_{\tdvecindex{\vecindex{s}{l}}{\vecindex{t}{j}}} B_{\vecindex{s}{l}}(X_s) B_{\vecindex{t}{j}}(X_t) + \displaystyle \sum_j \overline{B}_{\vecindex{t}{j}}(X_s)B_{\vecindex{t}{j}}(X_t) \\ = 
\displaystyle \sum_l \sum_j \overline{\theta}_{\tdvecindex{\vecindex{s}{l}}{\vecindex{t}{j}}} B_{\vecindex{s}{l}}(X_s) B_{\vecindex{t}{j}}(X_t) + \displaystyle \sum_l \overline{B}_{\vecindex{s}{l}}(X_t)B_{\vecindex{s}{l}}(X_s)
\end{split}
\end{equation}
Rearranging terms in the above equation gives us the following equation:
\begin{equation}
\label{vector_theorem_expanded_equality2}
\begin{array}{l}
\displaystyle \sum_j \left(\sum_l (\theta_{\tdvecindex{\vecindex{s}{l}}{\vecindex{t}{j}}}-\overline{\theta}_{\tdvecindex{\vecindex{s}{l}}{\vecindex{t}{j}}}) B_{\vecindex{s}{l}}(X_s) + \overline{B}_{\vecindex{t}{j}}(X_s)\right) B_{\vecindex{t}{j}}(X_t) \\ = 
 \displaystyle \sum_l B_{\vecindex{s}{l}}(X_s)\overline{B}_{\vecindex{s}{l}}(X_t)
\end{array}
\end{equation}

However, since $\forall l$ $\overline{B}_{\vecindex{s}{l}}(X_t)$ is orthogonal to $B_{t}(X_t)$, the left and right hand sides of the above equation are equal to 0, which leads us to the following equations.
\begin{equation}
\label{vector_theorem_expanded_equality3}
\begin{array}{l}
\displaystyle \sum_l B_{\vecindex{s}{l}}(X_s)\overline{B}_{\vecindex{s}{l}}(X_t) = 0\\
\\
\displaystyle \sum_j \left(\sum_l (\theta_{\tdvecindex{\vecindex{s}{l}}{\vecindex{t}{j}}}-\overline{\theta}_{\tdvecindex{\vecindex{s}{l}}{\vecindex{t}{j}}}) B_{\vecindex{s}{l}}(X_s) + \overline{B}_{\vecindex{t}{j}}(X_s)\right) B_{\vecindex{t}{j}}(X_t) = 0
\end{array}
\end{equation}
However since we assumed that the sufficient statistics are minimal we get $\forall l$ $\overline{B}_{\vecindex{s}{l}}(X_t) = 0$ from the first equality and $\forall j, l$ $\theta_{\tdvecindex{\vecindex{s}{l}}{\vecindex{t}{j}}} = \overline{\theta}_{\tdvecindex{\vecindex{s}{l}}{\vecindex{t}{j}}}$, $\overline{B}_{\vecindex{t}{j}}(X_s) = 0$ from the second equality.

Hence from \eqref{vector_theorem_basis_expansion2}, we obtain $E_s(X_t) = \theta_{st}(B_t(X_t) - B_t(0))$ and $E_t(X_s) = \theta_{st}^T(B_s(X_s) - B_s(0))$ where $\theta_{st}$ is a matrix formed by the scalars $\theta_{\tdvecindex{\vecindex{s}{l}}{\vecindex{t}{j}}}$ such that $(\theta_{st})_{lj} = \theta_{\tdvecindex{\vecindex{s}{l}}{\vecindex{t}{j}}}$ and:
\begin{equation}
\label{vector_theorem_tuple}
\begin{array}{l}
\mathcal{I}[X_s \neq 0, X_t \neq 0]G_{s,t}(X_s,X_t) = \\
(B_t(X_t) - B_t(0))^T\theta_{st}^T(B_s(X_s) - B_s(0))
\end{array}
\end{equation}
By extending this argument to higher order factors we can show that the natural parameters are required to be in the form specified by \eqref{vsmrf-vector_natural_parameters}.

\section{Proof of Sparsistency}
\label{sec:sparsistency_proof}

Before proving the sparsistency result, we will show that the sufficient statistics $B_{r}(X_r)$ are well behaved. Recall that   $B_{\vecindex{r}{i}}(X_r)$  indicates $i^{th}$ component of the vector   $B_{r}(X_r)$. We set the convention that whenever a variable has the subscript $\setminus r$ attached  we will be referring to the set of indexes $\{(t,j,k):\theta_{\tdvecindex{\vecindex{r}{j}}{\vecindex{t}{k}}} \in \theta_{r\cdot}, t\neq r \}$.


\begin{proposition}
\label{bound_on_sum_squares_of_B}
Let $\{X^{(j)}\}^{n}_{j=1}$ have joint distribution as in \eqref{vsmrf-vector_joint_distribution}, then, 
\begin{equation}
\label{bound_on_sum_squares}
P\left(\frac{1}{n}\underset{j=1}{\overset{n}{\sum}}\left(B_{\vecindex{r}{i}}\left(X_{r}^{\left(j\right)}\right)\right)^2\geq\delta\right) \leq \exp\left(-n\frac{\delta^{2}}{4k_{h}^{2}}\right)
\end{equation}
for  $\delta$ $\leq$ $\min\left\{ 2\frac{k_{v}}{3},k_{h}+k_{v}\right\} $.
\end{proposition}
\begin{proof}
It is clear from Taylor Series expansion and  assumption \ref{vsmrf-log_partition_joint_assumption} that
\begin{equation}
\label{eqn1}
\begin{array}{l}
\log E\left[\exp\left(tB_{\vecindex{r}{i}}\left(X_{r}\right)^2\right)\right]  = \\
 \log \int_{\otimes_{s\in [p]} \X_s} \exp \left\lbrace   t B_{\vecindex{r}{i}}\left(X_{r}\right)^{2}  + \right. \\
 \underset{s\in V}{\sum} \left\langle B_{s}\left(X_{s}\right), \theta_{s}^{*}  + \underset{t\in N(r)}{\sum}\theta_{st}^{*} B_{t} \left(X_{t} \right)\right\rangle  +\\
 \left. \underset{s\in V}{\sum}C_{s}\left(X_{s}\right) - A(\theta^*) \right\rbrace \mathit{v}\left(dx\right)\vspace{0.1in}\\
=\bar{A}_{r,i}\left(\eta;\theta\right)(t;\theta^{*}) - \bar{A}_{r,i}\left(\eta;\theta\right)(0;\theta^{*}) \vspace{0.1in}\\
  \leq  t \frac{\partial\bar{A}_{r,i}\left(\eta;\theta\right)}{\partial\eta}(0) + \frac{t^{2}}{2} \frac{\partial^{2}\bar{A}_{r,i}\left(\eta;\theta\right)}{\partial\eta^{2}}(ut) \\
   \leq  t\,k_v + \frac{t^{2}}{2} \,k_h 
\end{array}
\end{equation}
where $u \in [0,1]$

Therefore, by the standard Chernoff bounding technique, for $t \leq 1 $, it follows that
\begin{equation}
\label{eqn2}
\begin{array}{l}
\P \left(\frac{1}{n} \sum_{j=1}^{n} B_{\vecindex{r}{i}}\left(X_{r}^{\left(j\right)}\right)^2 \geq \delta\right) \leq \\
   \hspace{0.1in} \exp \left(
-n\delta t + n\,k_v t + \frac{t^{2}}{2}K_h n\right) \leq \\
 \hspace{0.1in}  \exp\left(-n\frac{\delta^{2}}{4k_{h}^{2}}\right) 
\end{array}
\end{equation}
for  $\delta$ $\leq$ $\min\left\{ 2\frac{k_{v}}{3},k_{h}+k_{v}\right\} $.
\end{proof}

\begin{proposition}
\label{bound_x_proposition}
Let X be a random vector with the distribution specified in \eqref{vsmrf-vector_joint_distribution}. Then, for any positive constant $\delta$ and some constant $c > 0$
\begin{equation}
P\Big(|B_{\vecindex{r}{i}}(X_r)| \geq \delta log(\eta)\Big) \leq c\eta^{-\delta}
\end{equation}
\end{proposition}
\begin{proof}
Let $\bar{v}$ be a unit vector with the same dimensions as $\theta_{r\cdot}^{*}$ and exactly one non-zero entry, corresponding to the sufficient statistic $B_{\vecindex{r}{i}}(X_r)$. Then we can write $\log \Big( E[\exp(B_{\vecindex{r}{i}}(X_r))]\Big)$ as:
\begin{equation*}
\log \Big(E[\exp(B_{\vecindex{r}{i}}(X_r))]\Big) = A(\theta^* + \bar{v}) - A(\theta^*)
\end{equation*} 
By Taylor series expansion, for some $u \in [0,1]$, we can rewrite last equation as
\begin{equation*}
\begin{split}
A(\theta^* + \bar{v}) - A(\theta^*)  = & \nabla A(\theta^*).\bar{v} + \frac{1}{2}\bar{v}^T\nabla^2 A(\theta^* + u\,\bar{v})\bar{v} \\
 =& E[B_{\vecindex{r}{i}}(X_r)]\|\bar{v}\|_2 \\
 & +  \frac{1}{2}\frac{\partial^2 A(\theta^* + u\,\bar{v})}{\partial \theta_{r_i}^2}\|\bar{v}\|_2^2 
\end{split}
\end{equation*}
Using Assumption \ref{vsmrf-log_partition_joint_assumption}  we get the inequality :
\begin{equation*}
A(\theta^* + \bar{v}) - A(\theta^*) \leq k_m + \frac{1}{2}k_h
\end{equation*}
Now, by using Chernoff bound, for any positive constant a, we get $P(B_{\vecindex{r}{i}}(X_r) \geq a) \leq \exp(-a + k_m + \frac{1}{2}k_h)$. By setting $a = \delta log(\eta)$ it follows that 
\begin{equation*}
P(B_{\vecindex{r}{i}}(X_r) \geq \delta log(\eta)) \leq \exp(-\delta log(\eta) + k_m + \frac{1}{2}k_h) \leq c\eta^{-\delta}
\end{equation*} where $c =\exp(k_m + \frac{1}{2}k_h)$ 
\end{proof}
The  proof of Sparsistency is based on the primal dual witness proof technique. 
First note that the optimality condition of \eqref{vsmrf-optimization_problem}, can be written as: 
\begin{equation}
\label{optimality_condition}
\nabla\ell(\hat{\theta}_{r\cdot};\mathcal{D})+\lambda_1 \sum_{t:r\neq t} \sqrt{\nu_{rt}} \hat{Z}_{1,rt} + \lambda_2 \hat{Z}_{2} = 0
\end{equation} 
where  $\hat{Z}_{1,rt}$  $\in$ $\partial\parallel\hat{\theta}_{rt}\parallel_{2}$,  $\hat{Z}_{2}$ $\in$ $\partial\parallel\hat{\theta}_{\setminus r}\parallel_{1}$  and we denote $\hat{Z}$  $=$ $\left(\hat{Z}_1,\hat{Z}_2\right)$, where $\hat{Z}_1 = \{\hat{Z}_{1,rt}\}_{t \in V\setminus r}$. And sub-gradients $\hat{Z}_1$, $\hat{Z}_2$ should satisfy the following conditions:
\begin{equation}
\label{dual_feasibility}
\begin{array}{lll}
\forall i \,\,\,(\hat{Z}_2)_{i}& =& sign\left((\hat{\theta}_{r\cdot})_{i}\right)\,\,\, \text{if} \,\,\,(\hat{\theta}_{r\cdot})_{i} \neq 0 \\
 && |(\hat{Z}_2)_{i}| \leq 1 \,\,\, \text{otherwise}\\
 \\
\forall t \,\,\, \hat{Z}_{1,rt}& =& \frac{\hat{\theta}_{rt}}{\vecnorm{\hat{\theta}_{rt}}{2}}\,\,\, \text{if} \,\,\, \hat{\theta}_{rt} \neq 0\\
&& \vecnorm{\hat{Z}_{1,rt}}{2} \leq 1 \,\,\, \text{otherwise}
\end{array}
\end{equation} 

Note that we can think of $\hat{Z}_1$ and $\hat{Z}_2$  as dual variables by appealing to Lagrangian theory. The next lemma shows that graph structure recovery is guaranteed if the dual is strictly feasible.
\begin{lemma}
\label{dual_witness_lemma}
Suppose that there exists a primal-dual pair  $\left(\hat{\theta}_{r\cdot},\hat{Z}\right)$ for \eqref{vsmrf-optimization_problem} such that $\vecnorm{\hat{Z}_{1,S^{c}}}{\infty,2}$ $<$ $1$ and $\vecnorm{\hat{Z}_{2, S^{c}}}{\infty}$ $<$ $1$. Then, any optimal solution $\tilde{\theta}_{r\cdot}$ must satisfy $\left(\tilde{\theta}_{r\cdot}\right)_{S^{c}}$ $=$ $0$. Moreover, if the Hessian sub-matrix $[\nabla^2\ell(\hat{\theta}_{r\cdot})]_{SS}$ is positive definite then $\hat{\theta}_{\setminus r}$ is the unique optimal solution.
\end{lemma}

\begin{proof}

First, note that by Cauchy$-$Schwarz's and Holder's inequalities
\begin{equation}
\label{inequalities}
\langle\hat{Z}_{1,rt},\tilde{\theta}_{rt}\rangle \leq \parallel\tilde{\theta}_{rt}\parallel_{2} \text{and}  \langle\hat{Z}_{2},\tilde{\theta}_{\setminus r}\rangle\leq\parallel\tilde{\theta}_{\setminus r}\parallel_{1}.
\end{equation}
But from  \eqref{optimality_condition} and the primal optimality of $\hat{\theta}_{r\cdot}$ and $\tilde{\theta}_{r\cdot}$  for \eqref{vsmrf-optimization_problem}, 
\begin{equation}
\label{dual_feasibility_proof}
\begin{array}{lll}
&\ell\left(\tilde{\theta}_{r\cdot}\right)+\sum_{t\neq r}\lambda_{1}\sqrt{\nu_{rt}}\langle\hat{Z}_{1,rt},\tilde{\theta}_{rt}\rangle+\lambda_{2}\langle\hat{Z}_{2},\tilde{\theta}_{\setminus r}\rangle  \\ 
&\geq  \underset{\theta}{\min{}}  \ell\left(\theta_{r\cdot}\right)+\sum_{t\neq r}\lambda_{1}\sqrt{\nu_{rt}}\langle\hat{Z}_{1,rt},\theta_{rt}\rangle+\lambda_{2}\langle\hat{Z}_{2},\theta_{\setminus r}\rangle \\  
& =  \ell\left(\hat{\theta}_{r\cdot}\right)+\sum_{t\neq r}\lambda_{1}\sqrt{\nu_{rt}}\langle\hat{Z}_{1,rt},\hat{\theta}_{rt}\rangle+\lambda_{2}\langle\hat{Z}_{2},\hat{\theta}_{\setminus r}\rangle \\
& =  \ell\left(\tilde{\theta}_{r\cdot}\right)+\sum_{t\neq r}\lambda_{1}\sqrt{\nu_{rt}} \vecnorm{\tilde{\theta}_{rt}}{2} + \vecnorm{\tilde{\theta}_{\setminus r}}{1}
\end{array}
\end{equation}

hence, combining with \eqref{inequalities} with \eqref{dual_feasibility_proof}  it follows that $\sum_{t\neq r}\lambda_{1}\sqrt{\nu_{rt}} \vecnorm{\tilde{\theta}_{rt}}{2} + \vecnorm{\tilde{\theta}_{\setminus r}}{1}$ $=$ $\sum_{t\neq r}\lambda_{1}\sqrt{\nu_{rt}} \vecnorm{\hat{\theta}_{rt}}{2} + \vecnorm{\hat{\theta}_{\setminus r}}{1}$ $=$ $\sum_{t\neq r}\lambda_{1}\sqrt{\nu_{rt}}\langle\hat{Z}_{1,rt},\hat{\theta}_{rt}\rangle+\lambda_{2}\langle\hat{Z}_{2},\hat{\theta}_{\setminus r}\rangle$.  The result follows.

If the Hessian sub-matrix is positive definite for the restricted problem then the problem is strictly convex and has a unique solution.
 \end{proof}

Based on the above lemma, we prove sparsistency theorem by constructing a primal-dual witness $(\hat{\theta}_{r\cdot}, \hat{Z})$ with the following steps:
\begin{enumerate}
\item Set $(\hat{\theta}_{r\cdot})_S = argmin_{\left(\left(\theta_{r\cdot}\right)_S,0\right)} {\ell\left(\left(\theta_{r\cdot}\right)_S; \mathcal{D}\right)}$ ${+ \lambda_1 \sum_{t \in S} \sqrt{\nu_{rt}} \vecnorm{\theta_{rt}}{2} + \lambda_2 \vecnorm{(\theta_{r\cdot})_S}{1}}$ 
\item For $t\in S$,  we define $\hat{Z}_{1,rt} = \frac{\theta_{rt}}{\|\theta_{rt}\|_2}$ and then construct $\hat{Z}_{2,S}$ by the stationary condition.
\item  Set $(\hat{\theta}_{r\cdot})_{S^c} = 0$
\item Set $\hat{Z}_{2,S^{c}}$ such that $\vecnorm{\hat{Z}_{2, S^{c}}}{\infty}$ $<$ $1$
\item Set $\hat{Z}_{1,S^{c}}$ such that condition \eqref{optimality_condition} is satisfied.
\item The final step consists of showing, that the following conditions are satisfied:
\begin{enumerate}
\item \textit{strict dual feasibility} : the condition in Lemma \ref{dual_witness_lemma} holds with high probability
\item \textit{correct neighbourhood recovery}: the primal-dual pair specifies the neighbourhood of r, with high probability
\end{enumerate} 
\end{enumerate}

We begin by proving some key lemmas that are key to our main theorem. The sub-gradient optimality condition \eqref{optimality_condition} can be rewritten as:
\begin{equation}
\label{rewritten_optimality_condition}
\nabla\ell(\hat{\theta}_{r\cdot};\mathcal{D})-\nabla\ell(\theta^{*}_{r\cdot};\mathcal{D})  =  W^{n} -\lambda_1 \sum_{t:r\neq t} \sqrt{\nu_{rt}} \hat{Z}_{1,rt} - \lambda_2 \hat{Z}_{2} 
\end{equation}
where $W^{n} = -\nabla\ell(\theta^{*}_{r\cdot};\mathcal{D}) $ and $\theta^{*}_{r\cdot}$ is the true model parameter. By applying mean-value theorem coordinate wise to \eqref{rewritten_optimality_condition}, we get:
\begin{equation}
\label{rewritten_optimality_condition2}
\begin{array}{lll}
\nabla^{2}\ell(\theta^{*}_{r\cdot};\mathcal{D})[\hat{\theta}_{r\cdot} - \theta^*_{r\cdot}] &=& W^{n} -\lambda_1 \sum_{t:r\neq t} \sqrt{\nu_{rt}} \hat{Z}_{1,rt}\\ &&- \lambda_2 \hat{Z}_{2} + R^n
\end{array}
\end{equation}
where $R^n$ is the remainder term after applying mean-value theorem:$R^n_{j} = [\nabla^{2}\ell(\theta^{*}_{r\cdot};\mathcal{D}) - \nabla^{2}\ell(\bar{\theta}^{j}_{r\cdot};\mathcal{D})]_j^T(\hat{\theta}_{r\cdot} - \theta^*_{r\cdot})$ for some $\bar{\theta}_{r\cdot}^j$ on the line between $\hat{\theta}_{r\cdot}$ and $\theta^*_{r\cdot}$, and with $[.]^T_j$ denoting the j-th row of matrix. The following lemma controls the score term $W^n$

\begin{lemma}
\label{w_bound}
 Recall  $\nu_{r\,\text{max}}$ $=$ $\underset{t}{\text{max}} \,\nu_{rt}$ , $\nu_{r\,\text{min}}$ $=$ $\underset{t}{\text{min}} \,\nu_{rt}$, $p^{\prime}$ $=$ $\text{max}(n,p)$. Assume that 
\begin{equation}
\begin{array}{lll}
 &\frac{8(2-\alpha)}{\alpha} \sqrt{k_1(n,p)\,k_4\, \frac{\nu_{r\,\text{max}} \text{log} (p\nu_{r\,\text{max}})} {n\nu_{r\,\text{min}}}} \leq \\ 
 & \lambda_{1} + \lambda_{2} \leq \frac{4(2-\alpha)\sqrt{\nu_{r\,\text{max}}}}{\alpha\sqrt{\nu_{r\,\text{min}}}}\,k_1(n,p)\,k_2(n,p)\,k_4
 \end{array}
\end{equation}

for some constant $k_4$ $\leq$ $\text{min}\left\{2\frac{k_v}{3},k_h+k_v\right\}$ and suppose also that  $n$ $\geq$ $\frac{8\,k_{h}^{2}}{k_{4}^{2}}$ $\,\text{log}\left(\sum_{t} m_t\right)$ then,
\begin{equation}
\label{bound_on_gradient}
\begin{array}{lll}
 P\left( \| W_{ \setminus r}^n  \|_{\infty,2}  > 
\frac{\alpha}{2-\alpha} \frac{\sqrt{\nu_{r\,\text{min}}}\,\left(\lambda_{1}+ \lambda_{2}\right)}{4}\right) \leq  \\
1 - c_1p^{\prime -3}\left(\sum_t m_t\right) - \exp{(-c_2n)} - \exp{(-c_3n)}
\end{array}
\end{equation}

\end{lemma}

\begin{proof}
Define  $W_{t}^n$ $=$  $-{\nabla_{\theta_{rt}} \ell\left(\theta_{r\cdot}^{*};\mathcal{D}\right)}$. Let $W_{t,jk}^n$ be the element in $W_{t}^n$ corresponding to parameter $\theta_{\tdvecindex{\vecindex{r}{j}}{\vecindex{t}{k}}}$. Note that  $W_{t,jk}^n$  $=$  $\frac{1}{n}$$\sum_{t=1}^{n}  V_{t,jk}^{i}$ where 
\begin{equation*}
\begin{array}{lll}
V_{t,jk}^{i}& = B_{\vecindex{r}{j}}\left(X_{r}^{\left(i\right)}\right)B_{\vecindex{t}{k}}\left(X_{t}^{\left(i\right)}\right) - \\ 
&\nabla_{\theta_{\tdvecindex{\vecindex{r}{j}}{\vecindex{t}{k}}}} A_{r}\left(\theta_{r}^{*}+ \sum_{s \in V \setminus r} \theta_{rs}^{*} B_{s}\left(X_{s}^{(i)}\right) \right) B_{\vecindex{t}{k}}\left(X_{t}^{\left(i\right)}\right)
\end{array}
\end{equation*}
so for   $t^{\prime}$ $\in $  $\mathbb{R}$

\begin{equation*}
\label{proof_part_1}
\begin{array}{lll}
E\left[\text{exp}\left( t^{\prime}V_{t,jk}^{i} \right)| X_{V\setminus r}^{(i)}\right ] =  \\
   \displaystyle \int_{X_r^{(i)}} \text{exp} \text{\bigg\{} 
t^{\prime}\text{\bigg[}  B_{\vecindex{r}{j}}\left(X_{r}^{(i)}\right) B_{\vecindex{t}{k}}\left(X_{t}^{(i)}\right)  \\
-  \nabla_{\theta_{\tdvecindex{\vecindex{r}{j}}{\vecindex{t}{k}}}} A_{r} \left(\theta_{r}^{*}+ \sum_{s \in V \setminus r} \theta_{rs}^{*} B_{s}\left(X_{s}^{(i)}\right)\right)  B_{\vecindex{t}{k}}\left(X_{t}^{\left(i\right)}\right)\text{\bigg]} \\
+ C\left(X_{r}^{(i)}\right) \, + \theta_{r}^{*} B_{r}\left(X_{r}^{(i)}\right) + \underset{s \in V \setminus r} {\sum} B_{r}\left(X_{r}^{(i)}\right)\theta_{rs} ^{*} B_{s}\left(X_{s}^{(i)}\right) \\ 
 - A_{r}\left(\theta_{r}^{*}+ \underset{s \in V \setminus r} {\sum} \theta_{rs} ^{*} B_{s}\left(X_{s}^{(i)}\right)\right)
\text{\bigg\}} dX_{r}\\
  =  \text{exp} \text{\bigg\{}  A_{r} \left(\theta_{r}^{*} + t^{\prime} B_{\vecindex{t}{k}}\left(X_{t}^{(i)}\right) + \underset{s \in V \setminus r} {\sum} \theta_{rs} ^{*} B_{s}\left(X_{s}^{(i)}\right)\right) \\ 
 -  A_{r} \left( \theta_{r}^{*} +
  \underset{s \in V \setminus r} {\sum} \theta_{rs} ^{*} B_{s}\left(X_{s}^{(i)}\right) \right) \\
  - \nabla_{\theta_{\tdvecindex{\vecindex{r}{j}}{\vecindex{t}{k}}}} A_{r} \left(\theta_{r}^{*} + \underset{s \in V \setminus r} {\sum} \theta_{rs} ^{*} B_{s}\left(X_{s}^{(i)}\right)\right) t^{\prime} B_{\vecindex{t}{k}}\left(X_{t}^{(i)}\right)  \text{\bigg\}}\\
  =   \text{exp} \text{\bigg\{} \frac{\nabla_{\theta_{\tdvecindex{\vecindex{r}{j}}{\vecindex{t}{k}}},\theta_{\tdvecindex{\vecindex{r}{j}}{\vecindex{t}{k}}}}^{2}A_{r}(c)}{2}  B_{\vecindex{t}{k}}\left(X_{t}^{(i)}\right)^{2} t^{\prime2} \text{\bigg\}}
\end{array}
\end{equation*}

where $c = \theta_{r}^{*} +  \sum_{s \neq r} \theta_{rs}^{*} B_s\left(X_{s}^{i}\right) + v_{1} t^{\prime} B_{\vecindex{t}{k}}\left(X_{t}^{(i)}\right)$   for some  $v_1$  $\in$ $[0,1]$. Therefore,

\begin{equation*}
\label{proof_part_2}
\begin{array}{lll}
 \frac{1}{n} \displaystyle \sum_{i=1}^{n} \text{log} E\left[\text{exp}\left( t^{\prime}V_{t,jk}^{i} \right)| X_{V\setminus r}^{(i)}\right]  = \\
   \frac{1}{n} \displaystyle \sum_{i=1}^{n} \frac{1}{2} \nabla_{\theta_{\tdvecindex{\vecindex{r}{j}}{\vecindex{t}{k}}},\theta_{\tdvecindex{\vecindex{r}{j}}{\vecindex{t}{k}}}}^{2}A_{r}(c) B_{\vecindex{t}{k}}\left(X_{t}^{(i)}\right)^{2} t^{\prime2}
\end{array}
\end{equation*}

Next lets define event $\varepsilon_1 = \text{\big\{} \max_{i,t} \|B_{t}\left(X_{t}^{\left(i\right)}\right) \|_{\infty} \leq 4\,\text{log}p^{\prime} \text{\big\}}$. Then, from Proposition \ref{bound_x_proposition} we get $P(\varepsilon_1^{c}) \leq c_1np^{\prime -4}\left(\sum_t m_t\right) \leq c_1p^{\prime -3}\left(\sum_t m_t\right) $.
If $t^{\prime} \leq k_2\left(n,p\right)$,  Assumption \ref{vsmrf-node_conditional_assumption} implies that
\begin{equation*}
\label{proof_part_3}
\begin{array}{lll}
\frac{1}{n} \displaystyle \sum_{i=1}^{n} \text{log} E\left[\text{exp}\left( t^{\prime}V_{t,jk}^{i} \right)| X_{V\setminus r}^{(i)}\right]   \leq \\
  \frac{k_{1}(n,p)}{2} \frac{1}{n} \sum_{i=1}^{n} B_{\vecindex{t}{k}}\left(X_{t}^{(i)}\right)^{2}t^{\prime2}
 \end{array}
\end{equation*}

%
%

Now, lets define event $\varepsilon_2 = \text{\big\{} \underset{t,j}{max} \frac{1}{n}  \sum_{i=1}^{n} \left(B_{\vecindex{t}{j}}\left(X_{t}^{(i)}\right)\right)^{2}
 \leq  k_4 \text{\big\}}$ where $k_4 \leq \min\{2k_v/3, k_h + k_v\}$.  Then,  by proposition (\ref{bound_on_sum_squares_of_B}) we obtain  that if  $n$  $\geq$  $\frac{8\,k_{h}^{2}}{k_{4}^{2}}$$\,\text{log}(\sum_{t \in V} m_t)$ :

 \begin{equation}
\label{part_4}
P\left(\varepsilon_{2}^{c}\right) \leq  \text{exp} \left(-n\frac{k_{4}^{2}}{4k_h^2}+ \text{log}\left(\sum_{t \in V} m_t\right) \right)  \leq \text{exp}\left(-n\,c_2\right)
\end{equation}

Therefore, for $t^{\prime}$ $\leq$ $k_{2}(n,p)$,

\begin{equation}
\label{part_5}
\frac{1}{n} \displaystyle \sum_{i=1}^{n} \text{log} E\left[\text{exp}\left( t^{\prime}V_{t,jk}^{i} \right)| X_{V\setminus r}^{(i)}\right]  \leq \frac{k_{1}(n,p)k_{4}{t^{\prime}}^{2}}{2} 
\end{equation}

Hence, by the standard Chernoff bound technique, for  $t^{\prime}$ $\leq$ $k_{2}(n,p)$

\begin{equation}
\label{part_6}
\begin{array}{lll}
P\left(
\frac{1}{n} \displaystyle \sum_{i=1}^{n} |V_{t,jk}^{i}| > \delta \mid \varepsilon_1, \varepsilon_2 
\right) \leq\\
  2 \text{exp}\left(n\left(\frac{k_{1}(n,p)k_4{t^{\prime}}^{2}}{2}-t^{\prime}\delta\right)\right) 
\end{array}
\end{equation}

Setting $t^{\prime}$ $=$ $\frac{\delta}{k_{1}(n,p)k_{4}}$, for  $\delta$ $\leq$ $k_{1}(n,p)\,k_{2}(n,p)\,k_4$, we arrive to:

\begin{equation}
\label{part_7}
P\left(
\frac{1}{n} \displaystyle \sum_{i=1}^{n} |V_{t,jk}^{i}| > \delta \mid \varepsilon_1, \varepsilon_2 
\right) \leq 2\, \text{exp}\left( \frac{-n\delta^{2}}{2k_{1}(n,p)k_{4}}\right)
\end{equation}
Supposing that $\frac{\alpha\,\sqrt{\nu_{r\,\text{min}}}}{2-\alpha}$$\,$$\frac{\lambda_1+ \lambda_{2}}{4\,\sqrt{m_r \,m_t}}$ $\leq$  $k_{1}(n,p)\,k_{2}(n,p)\,k_4$. It then follows that  $\delta$  $=$ $\frac{\alpha\,\sqrt{\nu_{r\,\text{min}}}}{2-\alpha}$$\,\frac{\lambda_{1}+ \lambda_{2}}{4\,\sqrt{m_r \,m_t}} $  satisfies 

\begin{equation}
\label{part_8}
\begin{array}{lll}
P\left(
\frac{1}{n} \displaystyle \sum_{i=1}^{n} |V_{t,jk}^{i}| > \frac{\alpha\,\sqrt{\nu_{r\,\text{min}}}}{2-\alpha} \frac{\lambda_{1}+ \lambda_{2}}{4\,\sqrt{m_r \,m_t}} \mid \varepsilon_1, \varepsilon_2 
\right) \leq \\
2\, \text{exp}\left( \frac{-\alpha^{2}}{(2-\alpha)^{2}} \frac{\nu_{r\,\text{min}}\,n\,\left(\lambda_{1}+ \lambda_{2}\right)^{2}}{32\,k_{1}(n,p)\,k_4\,m_r \,m_t} 
\right)
\end{array}
\end{equation}

Form which, we obtain the following using union bound

\begin{equation}
\label{part_9}
\begin{array}{lll} 
P\left( \| W_{t}^{n}\|_{2} > \frac{\alpha}{2-\alpha} \frac{\sqrt{\nu_{r\,\text{min}}}\,\left(\lambda_{1}+ \lambda_{2}\right)}{4 } \,| \varepsilon_1,\varepsilon_2  \right)  \leq \\
  P\left( \| W_{t}^{n}\|_{\infty} > \frac{\alpha}{2-\alpha} \frac{\sqrt{\nu_{r\,\text{min}}}\,\left(\lambda_{1}+ \lambda_{2}\right)}{4\,\sqrt{m_r \,m_t}} \,| \varepsilon_1,\varepsilon_2
\right) \\
 \leq  \, 2\text{exp}\left( \frac{-\alpha^{2}}{(2-\alpha)^{2}} \frac{\nu_{r\,\text{min}}\,n\,\left(\lambda_{1}+ \lambda_{2}\right)^{2}}{32\,k_{1}(n,p)k_4\,m_r \,m_t} + \text{log}\left(\nu_{rt}\right) 
\right)
\end{array}
\end{equation} 

 and hence,

\begin{equation}
\label{part_10}
\begin{array}{lll}
P\left( \| W^{n} \|_{\infty,2} > 
\frac{\alpha}{2-\alpha} \frac{\sqrt{\nu_{r\,\text{min}}}\,\left(\lambda_{1}+ \lambda_{2}\right)}{4 } \,| \varepsilon_1,\varepsilon_2
\right)   \leq \\
2\, \displaystyle \text{exp}\left( \frac{-\alpha^{2}}{(2-\alpha)^{2}} \frac{\nu_{r\,\text{min}}\,n\,\left(\lambda_{1}+ \lambda_{2}\right)^{2}}{32\,k_{1}(n,p)k_4\, \nu_{r\,\text{max}}} + \text{log}\left( \nu_{r\,\text{max}}  \right) + \text{log} p 
\right) 
\end{array}
\end{equation}

Finally for $\lambda_1+ \lambda_{2} \geq \frac{8(2-\alpha)}{\alpha} \sqrt{k_1(n,p)\,k_4\, \frac{\nu_{r\,\text{max}} \text{log} (p\nu_{r\,\text{max}})} {n\nu_{r\,\text{min}}}}$, we obtain

\begin{equation}
\label{part_11}
\begin{array}{lll}
P\left( \| W^{n} \|_{\infty,2} > 
\frac{\alpha}{2-\alpha} \frac{\sqrt{\nu_{r\,\text{min}}}\,\left(\lambda_{1}+ \lambda_{2}\right)}{4 }
\right)   \leq \\
c_1p^{\prime -3}\left(\sum_t m_t\right) + \exp{(-c_2n)} + \exp{(-c_3n)}
\end{array}
\end{equation}
\end{proof}

\begin{lemma}
\label{theta_bound}
Suppose that $\lambda_1+ \lambda_{2}$  $\leq$ $\frac{C_{\text{min}}^{2}}{40\,\text{log}p^{\prime}\,D_{\text{max}}\,d_r\,k_3\left(n,p\right)\,\nu_{r\,\text{max}}^{2}}$  and $\|W_{\setminus r}^{n}\|_{\infty,2}$ $\leq$ $\frac{\left(\lambda_1+ \lambda_{2}\right)\,\alpha\,\sqrt{\nu_{r\,\text{min}}}}{4\,\left(2-\alpha\right)}$, then,

\begin{equation}
\label{bound_on_theta}
\begin{array}{lll}
P\left( \|(\theta_{r\cdot}^{*})_{S}-(\hat{\theta}_{r\cdot})_{S}\|_{\infty,2} \leq 
\frac{5\,\sqrt{\nu_{r\text{max}}}}{C_{\text{min}}}\,\left(\lambda_1+ \lambda_{2}\right)
\right)  \\
\geq 1-cp^{\prime -3}\left(\sum_t m_t\right)
\end{array}
\end{equation}

for some constant $c$ $>$ $0$.
\end{lemma}

\begin{proof}
We define $F(u_S)$ as:

\begin{equation}
\begin{array}{lll}
F(u_S) =& \ell\left(\left(\theta_{r\cdot}^{*}\right)_{S}+u_S;\mathcal{D}\right) - \ell\left(\left(\theta_{r\cdot}^{*}\right)_{S};\mathcal{D}\right) \\
&+ \lambda_{1}\displaystyle \sum_{t \in N(r)} \sqrt{\nu_{rt}}\left(\|\theta_{rt}^{*}+u_{rt}\|_2 - \|\theta_{rt}^{*}\|_2\right)\\
&+ \lambda_{2} \left(\|\left(\theta_{r\cdot}^{*}\right)_{S}+u_S\|_1 - \|\left(\theta_{r\cdot}^{*}\right)_{S}\|_1\right)
\end{array}
\end{equation}

From the construction of $\hat{\theta}_{r\cdot}$ it is clear that  $\hat{u}_s$ $=$ $(\hat{\theta}_{r\cdot})_S$ $-$ $(\theta_{r\cdot}^{*})_S$ minimizes $F$. And since $F(0)$ $=$ $0$, we have $F\left(\hat{u}_s\right)$ $\leq$ $0$. We now show that for some $B > 0$ with $\vecnorm{u_S}{\infty, 2} = B$, we have $F(u_S) > 0$. Using this and the fact that $F$ is convex we can then show that $\vecnorm{\hat{u}_S}{\infty, 2} \leq B$.

Let $u_S$ an arbitrary vector   with $\vecnorm{u_S}{\infty, 2}$ $=$ $\frac{5\,\sqrt{\nu_{r\text{max}}}}{C_{\text{min}}}\,\left(\lambda_1 + \lambda_2\right)$. Then,  from  the Taylor Series expansion of log likelihood function in $F$, we have: 

\begin{equation}
\label{taylor_difference_bound}
\begin{array}{lll}
F(u_S)  =&  \nabla\ell\left(\left(\theta_{r\cdot}^{*}\right)_{S};\mathcal{D}\right)^{T} u_S  \vspace{0.05in} \\
& +u_S^{T} \nabla^{2}\ell\left(\left(\theta_{r\cdot}^{*}\right)_{S} + v\, u_S\right)\,u_S \\
&+ \lambda_{1}\displaystyle \sum_{t \in N(r)} \sqrt{\nu_{rt}}\left(\|\theta_{rt}^{*}+u_{rt}\|_2 - \|\theta_{rt}^{*}\|_2\right)
\\
 & +  \lambda_{2} \left(\|\left(\theta_{r\cdot}^{*}\right)_{S}+u_S\|_1 - \|\left(\theta_{r\cdot}^{*}\right)_{S}\|_1\right)
\end{array}
\end{equation}
for some $v \in [0,1]$

We now bound each of the terms in the right hand side of (\ref{taylor_difference_bound}). From \eqref{part_9} and using Cauchy-Schwarz inequality we obtain:

\begin{equation}
\begin{array}{lll}
 \left| \nabla\ell\left(\left(\theta_{r\cdot}^{*}\right)_S;\mathcal{D}\right)^{T} u_S \right| \\
  \leq   \|\nabla\ell\left(\left(\theta_{r\cdot}^{*}\right)_S;\mathcal{D}\right) \|_{\infty} \, \|u_S\|_{1} \\
  \leq   \|\nabla\ell\left(\left(\theta_{r\cdot}^{*}\right)_S;\mathcal{D}\right) \|_{\infty} \,d_r \,\sqrt{\nu_{r\,\text{max}}}\, \|u_S\|_{\infty,2} \\
 \leq   \frac{\alpha}{2-\alpha} \frac{\lambda_{1}+\lambda_2}{4} \,d_r \,\nu_{r\,\text{max}}\,\frac{5}{C_{\text{min}}}\,\left(\lambda_1+\lambda_2\right)\\
 =  \frac{5\,\nu_{r\,\text{max}}}{4\,C_{\text{min}}}\,d_r\,\left(\lambda_1+\lambda_2\right)^{2}
\end{array}
\end{equation}

where the last inequality holds because $\alpha \in (0,1]$.
Moreover, from triangle inequality we have:
\begin{equation}
\begin{array}{lll}
\lambda_1\displaystyle \sum_{t \in S} \sqrt{\nu_{rt}}\left(\|\theta_{rt}^{*}+u_{rt}\|_2 - \|\theta_{rt}^{*}\|_2\right) \\
 \geq   -\lambda_1 \displaystyle \sum_{t \in N(r)} \sqrt{\nu_{rt}} \|u_{rt}\|_{2} \\
 \geq  -\lambda_1 \,d_r\,\sqrt{\nu_{r\,\text{max}}}\, \|u_S\|_{\infty,2} \\
 =   -\frac{5\,d_r\,\nu_{r\,\text{max}}}{C_{\text{min}}}\,\lambda_{1}\left(\lambda_{1}+\lambda_2\right)
\end{array}
\end{equation}

Also,
\begin{equation}
\begin{array}{lll}
\lambda_2\,\left(\|\left(\theta_{r\cdot}^{*}\right)_S+u_S\|_1 - \|\left(\theta_{r\cdot}^{*}\right)_S\|_1\right)  \\
\geq  -\lambda_2\,\|u_S\|_{1} \\
 \geq  -\lambda_2\,\|u_S\|_{1} \\
 \geq  -\lambda_2\,d_r\,\sqrt{\nu_{r\,\text{max}}}\,\|u_S\|_{\infty,2} \\
  =   -\frac{5\,\nu_{r\,\text{max}}}{C_{\text{min}}}\,d_r\,\lambda_{2}\left(\lambda_{1}+\lambda_2\right)
\end{array}
\end{equation}

On the other hand, by Taylor's approximation of $\nabla^{2}\ell$, there exists $\alpha_{jk}$ $\in$ $[0,1]$  and $\tilde{u}_{jk}^{i}$ between $\theta_{\setminus r}$ and $\theta_{\setminus r}+vu_S$  such that 

\begin{equation}
\begin{array}{lll}
 \Lambda_{\text{min}}\left(\nabla^{2}\ell\left(\left(\theta_{r\cdot}^{*}\right)_S + v\,u_S\right)\right) \\
  \geq  \underset{\beta \in [0,1]}{\text{min}}  \Lambda_{\text{min}}\left(\nabla^{2}\ell\left(\left(\theta_{r\cdot}^{*}\right)_S + \beta\,u_S\right)\right) \\
  \geq  \Lambda_{\text{min}}\left(Q_{SS}^{n}\right) - \\
  \underset{v \in [0,1]}{\text{max}} \underset{ \|y\| \leq 1}{\text{max}}\Bigg\{ \frac{1}{n} \displaystyle \sum_{i=1}^{n} \displaystyle \sum_{j,k,l,t,h,s,m,s\prime,m\prime} \alpha_{jk}\,v\,\left(\nabla^{3}A_r\left(\tilde{u}_{jk}^{i}\right)\right)_{jkl}\\
  u_{rt;lh}\,B_{\vecindex{t}{h}}\left(X_{t}^{i}\right)\,y_{s,m,j}\,B_{\vecindex{s}{m}}\left(X_{s}^{i}\right)\,B_{\vecindex{s^{\prime}}{m^{\prime}}}\left(X_{s^{\prime}}^{i}\right)\,y_{s^{\prime},m^{\prime},j^{\prime}}
 \Bigg\}
 \end{array}
\end{equation}
Consider the event $ \varepsilon_1$ as defined in the previous proof. We know that $P(\varepsilon_1) \geq 1 - c_1p^{\prime -3}\left(\sum_t m_t\right)$. Conditioned on $\varepsilon_1$ and using Assumption \ref{vsmrf-node_conditional_assumption} we arrive to the following:
\begin{equation}
\begin{array}{lll}
\Lambda_{\text{min}}\left(\nabla^{2}\ell\left(\left(\theta_{r\cdot}^{*}\right)_S + v\,u_S\right)\right) \\
  \geq  C_{\text{min}} - 4\text{log}p^{\prime}\|u_S\|_1\,D_{\text{max}}\,\nu_{r\,\text{max}}k_3\left(n,p\right) \\ 
  \geq   C_{\text{min}} - 4\text{log}p^{\prime}d_r
 \sqrt{\nu_{r\,\text{max}}}\vecnorm{u_S}{\infty,2}\,
 D_{\text{max}}\,\nu_{r\,\text{max}}k_3\left(n,p\right) \\
  \geq \frac{C_{\text{min}}}{2}
\end{array}
\end{equation}
where the last inequality holds for  $\lambda_1 + \lambda_2$  $\leq$ $\frac{C_{\text{min}}^{2}}{40\,\text{log}p^{\prime}\,D_{\text{max}}\,d_r\,k_3\left(n,p\right)\,\nu_{r\,\text{max}}^{2}}$.

Finally using the above bounds we arrive at the following:
\begin{equation}
F\left(u_S\right) \geq d_r\,\nu_{r\,\text{max}}\,\frac{5}{C_{\text{min}}}\,\left(\lambda_{1}+\lambda_2\right)^{2}\,\left( -1 -\frac{1}{4} + \frac{5}{2} 
\right) > 0
\end{equation}

Therefore
\begin{equation*}
\|(\theta_{r\cdot}^{*})_{S}-(\hat{\theta}_{r\cdot})_{S}\|_{\infty,2} \leq 
\frac{5\,\sqrt{\nu_{r\text{max}}}}{C_{\text{min}}}\,\left(\lambda_1+\lambda_2\right)
\end{equation*}
\end{proof}

\begin{lemma}
\label{r_bound}
Suppose that $\lambda_1 + \lambda_2$ $\leq $  $\frac{\alpha}{2-\alpha} \frac{\sqrt{\nu_{r\,\text{min}}}\,C_{\text{min}}^{2}}{400\,\nu_{r\,\text{max}}^{\frac{5}{2}} \text{log}p^{\prime}\,D_{\text{max}} \,k_3\left(n,p\right)\,d_r }$  and $\|W_{\setminus r}^{n}\|_{\infty,2}$ $\leq$ $\frac{\alpha\,(\lambda_1 + \lambda_1)\,\sqrt{\nu_{r\,\text{min}}}}{4\,\left(2-\alpha\right)}$, then,

\begin{equation}
\label{bound_on_R}
P\left( \frac{\|R^{n}\|_{\infty,2}}{\lambda_1 + \lambda_2} \leq \frac{\alpha\,\sqrt{\nu_{r\,\text{min}}}}{4\left(2-\alpha\right)}
\right)  \geq 1-cp^{\prime -3}\left(\sum_t m_t\right)
\end{equation}

for some constant $c$ $>$ $0$.
\end{lemma}

\begin{proof}
Recall that   $R_{j}^{n}$ $=$ $\big[ \nabla^{2}\ell(\theta^{*}_{r\cdot};\mathcal{D})- \nabla^{2}\ell(\bar{\theta}^{j}_{r\cdot};\mathcal{D}) \big]_{j}^{T} \left(\hat{\theta}_{r\cdot}-\theta^{*}_{r\cdot}\right)$  where $\big[ \cdot \big]_{j}^{T}$  denotes the j-th row of a matrix. Let us also  refer to   $R_{t;jk}^{n}$ to the cordinate of $R^{n}$  corresponding to  $\theta_{\tdvecindex{\vecindex{r}{j}}{\vecindex{t}{k}}}$. Then,

\begin{equation}
\label{Rn_1}
\begin{array}{lll} 
R_{t;jk}^{n} = \Bigg[ \frac{1}{n} \displaystyle \sum_{i=1}^{n} B_{\vecindex{t}{k}}(X_{t}^{i})\Bigg[ \nabla^{2}A_{r}\left(\theta_{r}^{*}+\sum_{s \neq r} \theta_{rs}^{*}B_{s}(X_{s}^{i})\right)  - \\ \nabla^{2}A_{r}\left(\bar{\theta}_{r}^{t_{jk}}+\sum_{s \neq r} \bar{\theta}_{rs}^{t_{jk}}B_{s}(X_{s}^{i})\right) \Bigg]_{j} \otimes B_{r\cdot}^{i}\Bigg]^{T} \left(\hat{\theta}_{r\cdot}-\theta_{r\cdot}^{*}\right)
\end{array}
\end{equation} 

with  $B_{r\cdot}^{i}$  the vector of sufficient statistics evluate at the i-th  sample. Intoroducing the notation $\langle \theta_{r\cdot}, B_{r\cdot}^{i} \rangle$  $=:$ $\theta_r+ \sum_{s \neq r} \theta_{rs} B_s\left(X_{s}^{i}\right)$, from the mean value theorem we obtain 

\begin{equation}
\label{Rn_2}
\begin{array}{lll} 
\nabla^{2}A_{r;jl}\left(\langle \theta_{r\cdot}^{*}, B_{r\cdot}^{i} \rangle\right)  -  \nabla^{2}A_{r;jl}\left( \langle \bar{\theta}_{r\cdot}^{t_{jk}},B_{r\cdot}^{i} \rangle \right) 
=\\
 -v_{j,l}^{i} \big[  \langle \hat{\theta}_{r\cdot} -\theta_{r\cdot}^{*}, B_{r\cdot}^{i} \rangle \big] \left(\nabla^{3}A_{r} \right)_{jl:}\left(\langle \bar{\bar{\theta}}_{r\cdot}^{i;t_{jk}}, B_{r\cdot}^{i} \rangle\right)
\end{array}
\end{equation} 
Therefore, combining (\ref{Rn_1}) with (\ref{Rn_2}) and using basic properties of krockner product we ontain that   
\begin{equation}
\begin{array}{lll} 
\left|R_{t;jk}^{n}\right| \hspace{-0.1in}& \leq &\hspace{-0.1in} \frac{1}{n} \displaystyle\sum_{i=1}^{n} \left|B_{\vecindex{t}{k}}(X_{t}^{i})\right| \nu_{r\,\text{max}}k_{3}\left(n,p\right) \,D_{max}\,\|\hat{\theta}_{r\cdot}-\theta_{r\cdot}^{*}\|_{2}^{2}
\\
 & \leq &  4\,\text{log} p^{\prime}\, \nu_{r\,\text{max}}k_{3}\left(n,p\right) \,D_{max}\,\|\hat{\theta}_{r\cdot}-\theta_{r\cdot}^{*}\|_{2}^{2}
\end{array}
\end{equation} 
which implies

\begin{equation}
\begin{array}{lll} 
\|R_{t}^{n}\|_{\infty,2}  \leq  \\
4\, \sqrt{\nu_{r\,\text{max}}}\,\text{log} p^{\prime}\, \nu_{r\,\text{max}}k_{3}\left(n,p\right) \,D_{max}\,\|\hat{\theta}_{r\cdot}-\theta_{r\cdot}^{*}\|_{2}^{2} \leq\\
    4\, \sqrt{\nu_{r\,\text{max}}}\,\text{log} p^{\prime}\, \nu_{r\,\text{max}}k_{3}\left(n,p\right) \,D_{max}\,
\frac{25\,d_r\,\nu_{r\text{max}}}{C_{\text{min}}^2}\,\lambda_{1}^{2}\ \leq\\
   \frac{\lambda_1 + \lambda_2}{4}\,\frac{\alpha\,\sqrt{\nu_{r\,\text{min}}}}{2-\alpha}\, 
\end{array}
\end{equation}

with probbility at least $1-cp^{\prime -3}\left(\sum_t m_t\right)$.  
\end{proof}

We now prove theorem \ref{vsmrf-sparsistency_theorem_statement} using lemmas \ref{w_bound}-\ref{r_bound}. Recalling that $Q^n = \nabla^{2}\ell(\theta^{*}_{r\cdot};\mathcal{D})$ and the fact that we have set $(\hat{\theta}_{r\cdot})_{S^c} = 0$ in our primal-dual construction, we can rewrite condition \eqref{rewritten_optimality_condition2} as the following equations:
\begin{equation}
\label{rewritten_optimality_condition3}
\begin{array}{lll}
Q^n_{S^cS}[(\hat{\theta}_{r\cdot})_S - (\theta^*_{r\cdot})_S] = \\ W^n_{S^c} -\lambda_1 \sum_{t \notin N(r)} \sqrt{\nu_{rt}} \hat{Z}_{1,rt} - \lambda_2 \hat{Z}_{2,S^c} + R^n_{S^c}
\end{array}
\end{equation}

\begin{equation}
\label{rewritten_optimality_condition4}
\begin{array}{lll}
Q^n_{SS}[(\hat{\theta}_{r\cdot})_S - (\theta^*_{r\cdot})_S] = \\ W^n_{S} -\lambda_1 \sum_{t \in N(r)} \sqrt{\nu_{rt}} \hat{Z}_{1,rt} - \lambda_2 \hat{Z}_{2, S} + R^n_{S}
\end{array}
\end{equation}

Since the matrix $Q^n_{SS}$ is invertible, the conditions \eqref{rewritten_optimality_condition3} and \eqref{rewritten_optimality_condition4} can be rewritten as :
\begin{equation}
\begin{array}{lll}
Q^n_{S^cS}(Q^n_{SS})^{-1}[W^n_{S} -\lambda_1 \hspace{-0.1in}\displaystyle \sum_{t \in N(r)} \hspace{-0.1in}\sqrt{\nu_{rt}} \hat{Z}_{1,rt} - \lambda_2 \hat{Z}_{2, S} + R^n_{S}]=\\
W^n_{S^c} -\lambda_1 \sum_{t \notin N(r)} \sqrt{\nu_{rt}} \hat{Z}_{1,rt} - \lambda_2 \hat{Z}_{2,S^c} + R^n_{S^c}
\end{array}
\end{equation}
Rearranging yields the following condition:
\begin{equation}
\label{rewritten_optimality_condition5}
\begin{array}{lll}
\lambda_1 \sum_{t \notin N(r)} \sqrt{\nu_{rt}} \hat{Z}_{1,rt}  =\\
 W^n_{S^c} + R^n_{S^c} - Q^n_{S^cS}(Q^n_{SS})^{-1}[W^n_{S}  +
  R^n_{S}] -\\
   \lambda_2 \hat{Z}_{2,S^c} + Q^n_{S^cS}(Q^n_{SS})^{-1}[\lambda_1 \sum_{t \in N(r)} \sqrt{\nu_{rt}} \hat{Z}_{1,rt} + \lambda_2 \hat{Z}_{2, S}]
\end{array}
\end{equation}

\textit{Strict Dual Feasibility}: we now show that for the dual sub-vector $\hat{Z}_{1,S^c}$, we have $\vecnorm{\hat{Z}_{1,S^c}}{\infty, 2} < 1$. We get the following equation from  \ref{rewritten_optimality_condition5}, by applying triangle inequality:

\begin{equation}
\begin{array}{lll}
\lambda_1 \sqrt{\nu_{r\,\text{min}}} \vecnorm{\hat{Z}_{1,S^c}}{\infty, 2}  \leq \\ \lambda_1 \vecnorm{\sum_{t \notin N(r)} \sqrt{\nu_{rt}} \hat{Z}_{1,rt}}{\infty,2}   \leq \\
  \left[\vecnorm{W^n}{\infty, 2} + \vecnorm{R^n}{\infty, 2} \right] \left(1 + \vecnorm{Q^n_{S^cS}(Q^n_{SS})^{-1}}{\infty, 2} \sqrt{d_r}\right) \\
 + \lambda_2 \sqrt{\nu_{r\,\text{max}}} \\
 + \vecnorm{Q^n_{S^cS}(Q^n_{SS})^{-1}}{\infty, 2}\left[\left(\lambda_1 +\lambda_2\right) \sqrt{d_r\,\nu_{r\,\text{max}}}\right]
\end{array}
\end{equation}

where $\nu_{r\,\text{min}} $ $=$ $\underset{t}{\text{min}} \,\nu_{rt} \text{ , }$ $\nu_{r\,\text{max}} $ $=$ $\underset{t}{\text{max}} \,\nu_{rt}$ and $d_r = |N(r)|$
Using mutual incoherence bound \ref{vsmrf-mutual_incoherence_bound} on the above equation gives us:
\begin{equation}
\begin{array}{lll}
\vecnorm{\hat{Z}_{1,S^c}}{\infty, 2}  \leq  \\ 
\frac{1}{ \lambda_1 \sqrt{\nu_{r\,\text{min}}}}\left[\vecnorm{W^n}{\infty, 2} + \vecnorm{R^n}{\infty, 2} \right] \left(2-\alpha \right) \\
 + \frac{\lambda_2 \sqrt{\nu_{r\,\text{max}}}}{\lambda_1 \sqrt{\nu_{r\,\text{min}}}} \left[1 + \vecnorm{Q^n_{S^cS}(Q^n_{SS})^{-1}}{\infty, 2}\sqrt{d_r}\left(\frac{\lambda_1}{\lambda_2} + 1\right)\right] \\
\end{array}
\end{equation}

Using the previous lemmas we obtain the following:
\begin{equation}
\begin{array}{lll}
\vecnorm{\hat{Z}_{1,S^c}}{\infty, 2}  \leq   
\frac{1}{ 2\lambda_1}\left[\alpha \left(\lambda_1 + \lambda_2\right) \right] \\
 + \frac{\lambda_2 \sqrt{\nu_{r\,\text{max}}}}{\lambda_1 \sqrt{\nu_{r\,\text{min}}}} \left[1 + \frac{m_{min}}{m_{max}}(1-\alpha)\left(\frac{\lambda_1}{\lambda_2} + 1\right)\right] \\
 \end{array}
\end{equation}

If  $\lambda_2$ $<$ $\left(\frac{\alpha}{2 - \alpha + 2 \frac{\sqrt{\nu_{r\,\text{max}}}}{\sqrt{\nu_{r\,\text{min}}}}}\right)\, \lambda_1$, then,
\begin{equation}
\begin{array}{lll}
\vecnorm{\hat{Z}_{1,S^c}}{\infty, 2} & < &  1
\end{array}
\end{equation}
We have shown that the dual is strictly feasible with high probability and also the solution is unique. And hence based on Lemma \ref{dual_witness_lemma} the method correctly excludes all edges not in the set of edges.

\textit{Correct Neighbourhood Recovery}: To show that all correct neighbours are recovered, it suffices to show that
\begin{equation*}
\|(\theta_{r\cdot}^{*})_{S}-(\hat{\theta}_{r\cdot})_{S}\|_{\infty,2} \leq \frac{\theta_{min}}{2}
\end{equation*}
where $\theta_{min} = \min_{t \in V\setminus r}\vecnorm{\theta_{rt}}{2}$.

Using Lemma \ref{theta_bound} we can show the above inequality holds if $\theta_{min} \geq \frac{10\ \sqrt{\nu_{r\,max}}}{C_{\text{min}}}\left(\lambda_{1}+\lambda_{2}\right)$

\section{Full MyFitnessPal Graph}
\label{sec:full_mfp_graph}

\begin{figure*}[tbh]
\begin{center}
\includegraphics[trim = 0in 1.5in 0in 1.5in, scale=0.5]{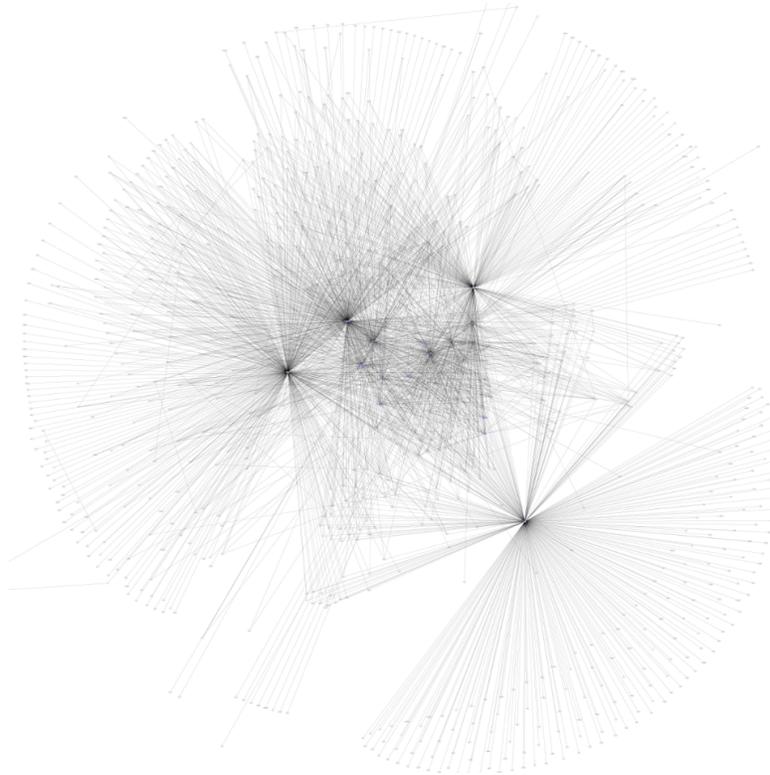}
\end{center}
\caption{\label{fig:mfp_full_graph} Full MRF learned from the MyFitnessPal food database. The hubs correspond to point-inflated gamma nutrient nodes, with the three largest hubs being the macro-nutrients (fat, carbs, and protein).}
\end{figure*}

Figure \ref{fig:mfp_full_graph} shows a high-level view of the entire VS-MRF learned from the MyFitnessPal food database. The three macro-nutrients (fat, carbs, and protein) correspond to the three largest hubs with the remaining nine micro-nutrients representing smaller hubs.

\bibliographystyle{abbrvnat}
\bibliography{preprint}

\end{document}